\documentclass[preprint,12pt]{elsarticle}

\usepackage{amssymb}
\usepackage{amsmath}
\usepackage{float}
\usepackage[unicode]{hyperref}
\usepackage{xcolor}
\usepackage{ulem}
\usepackage{graphicx}
\begin{document}

\begin{frontmatter}

\title{Process-Informed Forecasting of Complex Thermal Dynamics in Pharmaceutical Manufacturing}

\author[label1,label2,label3]{Ramona Rubini \corref{cor1}} %% Author name
\author[label3]{Siavash Khodakarami} %% Author name
\ead{siavash_khodakarami@brown.edu}
\author[label3]{Aniruddha Bora} %% Author name
\ead{aniruddha_bora@brown.edu}
\author[label3]{George Em Karniadakis} %% Author name
\ead{george_karniadakis@brown.edu}
\author[label2]{Michele Dassisti} %% Author name
\ead{michele.dassisti@poliba.it}

\cortext[cor1]{ramona.rubini@uniba.it, ramona\_rubini@brown.edu}
\affiliation[label1]{organization={Department of Agricultural and Environmental Sciences, University of Bari},%Department and Organization
            city={Bari},
            postcode={70121}, 
            country={Italy}}
            
\affiliation[label2]{organization={Department of Mechanical, Mathematics, and Management (DMMM), Polytechnic University of Bari},%Department and Organization
            city={Bari},
            postcode={70126}, 
            country={Italy}}

\affiliation[label3]{organization={Division of Applied Mathematics, Brown University},%Department and Organization
            city={Providence},
            postcode={02912}, 
            state={RI},
            country={USA}}

%% Abstract
\begin{abstract}
Accurate time-series forecasting for complex physical systems is the backbone of modern industrial monitoring and control, yet deep learning models often lack the physical consistency required in regulated environments.To bridge this gap, we introduce Process-Informed Forecasting (PIF) models for temperature in pharmaceutical lyophilization, embedding deterministic production recipes as macro-structural priors. We investigate classical methods (e.g., Autoregressive Integrated Moving Average (ARIMA) model) and modern deep learning architectures, including Kolmogorov-Arnold Networks (KANs). We compare three different loss function formulations that integrate a process-informed trajectory prior: a fixed-weight loss, a dynamic uncertainty-based loss, and a Residual-Based Attention (RBA) mechanism. We evaluate all models not only for accuracy and physical consistency but also for robustness to sensor noise. Furthermore, we test the practical generalizability of the best model in a transfer-learning scenario to a new process. Our results show that PIF models outperform their data-driven counterparts in terms of accuracy, physical plausibility and noise resilience, offering a scalable framework for reliable and generalizable forecasting solutions in critical manufacturing.
\end{abstract}

% \begin{highlights}
%     \item We formulate an idealized, process-informed trajectory prior and compare three different loss function formulations for deep learning models: a fixed-weight loss, a dynamic uncertainty-based loss and a Residual-Based Attention (RBA) mechanism. 
    
%     \item We compare process-informed forecasting models with a wide range of traditional and state-of-the-art data-driven counterparts. To ensure fairness and clarity, all deep learning architectures are calibrated to the same parameter count. Our evaluation assesses not only predictive accuracy but also physical plausibility and robustness to sensor noise.
    
%     \item Moreover, we validate the practical utility of our methodology through a transfer learning study, assessing the generalizability of the best-performing model on a new, unseen temperature dynamics.
% \end{highlights}

%% Keywords

\end{frontmatter}



\section{Introduction}
\label{sec1}
In pharmaceutical manufacturing, robust time-series forecasting is key to ensuring product quality, optimizing processes, and enabling predictive maintenance to prevent costly production disruptions. A particularly critical process in this industry is lyophilization, or freeze-drying, used to preserve and stabilize sensitive pharmaceutical products such as vaccines, antibiotics, and biologics. In this process, the product's moisture is removed at low temperature via sublimation \cite{Nail2002}. Given the thermal sensitivity of pharmaceutical products, accurate prediction and control of product temperature during lyophilization are critical \cite{rubini2023a, rubini2023b}. 
Classical statistical methods like Autoregressive Integrated Moving Average (ARIMA) and Exponential Smoothing (ETS) have been the benchmark for forecasting applications, valued for their simplicity, interpretability, and effectiveness in linear or trend-based data \cite{Box_2015, storti2018, gardner1985, pegels1969, wang2011, korzenowski2013, muhr2021, kalman1960, jeffries2002}. 

However, classical approaches are no longer sufficient to capture the complex nonlinear thermal dynamics of physical systems in real-world contexts \cite{sowemimo2024, cui2024, oukassi2023}. This data-rich environment has triggered a paradigm shift to deep learning. In particular, Long Short-Term Memory (LSTM) and the more recent attention-based Transformers have emerged as powerful alternatives to model temporal dependencies and often outperform classical methods \cite{el2022, Han_2024, alsaedi2025, abbasimehr2020, elalem2023, dutta2021, varghese2024}. While highly effective for long-range dependencies, standard data-driven models often ignore the underlying physics of the processes. For instance, the global self-attention mechanism of standard Transformers can sometimes act as a smoothing filter, making it challenging to capture sharp, high-frequency local transients in strict thermodynamic sequences without specialized tuning \cite{zeng2023}. This lack of interpretability can compromise their generalizability, data efficiency and reliability, especially in the pharmaceutical industry, which is subject to strict Good x Practice (GxP) regulations.

The integration of deep learning into critical sectors extends far beyond simple forecasting. Recent advancements have demonstrated the power of enhanced architectures, such as Recurrent Neural Networks (RNNs) optimized with hybrid algorithms for complex motion tracking and U-Nets with attention mechanisms for precise medical image segmentation \cite{cheltha2024enhanced, aung2025enhanced}. However, deploying such data-driven intelligence in the pharmaceutical industry—whether for optimizing lyophilization or for revolutionizing vaccine supply chains via federated blockchain frameworks—raises significant concerns about the security of proprietary data \cite{saha2025revolutionising}. As manufacturing transitions towards Industry 4.0, ensuring privacy-aware secure data auditing for cloud-based intelligence and exploring decentralized frameworks becomes paramount to comply with industrial regulations \cite{ullah2025privacy, khan2023privacy}.

To bridge the gap between data-driven power and scientific rigor, recent advances have introduced Physics-Informed Neural Networks (PINNs), which incorporate physical knowledge directly into the learning process \cite{Raissi_2019}. 
Traditional PINNs rely on embedding complex, continuous Partial Differential Equations (PDEs) into the loss function. However, in pharmaceutical lyophilization, deriving accurate PDEs is often impractical due to unknown, product-specific thermodynamic parameters (e.g., vial heat transfer coefficients or cake resistance to mass flow). To overcome this, we propose a shift from Equation-Informed to Process-Informed forecasting. Instead of relying on theoretical physics equations, our integration guides the model using the deterministic Production Recipe, which is the actual machine instructions programmed into the Programmable Logic Controller (PLC), as a macro-structural prior.
This integration guides the model towards physically consistent solutions, even in the presence of sparse or noisy input, therefore improving its understanding and robustness. A further innovation has come through the development of Kolmogorov-Arnold Networks (KANs) \cite{liu_2025}.  Inspired by the Kolmogorov-Arnold representation theorem, KANs have learnable univariate activation functions on the edges of the network that improve interpretability and parameter efficiency compared to traditional Multi-Layer Perceptrons (MLPs) \cite{vaca2024, pourkamali2024}. Extensions such as the Chebyshev polynomial-based (cKAN) and models like Long Expressive Memory (LEM) address additional limitations, such as spectral bias, that can lead to inaccurate modeling of high-frequency dynamics \cite{Sidharth_2024, rusch2021, kapoor_2024a}.

Building upon these advancements, this paper introduces Process-Informed Forecasting (PIF) models for pharmaceutical freeze-drying. Our main contributions can be summarized as follows: 
\begin{itemize}

    \item We formulate an idealized, process-informed trajectory prior and compare three different loss function formulations for deep learning models: a fixed-weight loss, a dynamic uncertainty-based loss \linebreak[0] \cite{Kendall_2018} and a Residual-Based Attention (RBA) mechanism \linebreak[0] \cite{anagnostopoulos2024}. 
    
    \item We compare PIF models with a wide range of traditional and state-of-the-art data-driven counterparts. To ensure fairness and clarity, all deep learning architectures are calibrated to the same parameter count. Our evaluation assesses not only predictive accuracy but also physical plausibility and robustness to sensor noise.
    
    \item Moreover, we validate the practical utility of our methodology through a transfer learning study, assessing the generalizability of the best-performing model on a new, unseen set of temperature dynamics.

    \item Our empirical results demonstrate that PIF models consistently outperform purely data-driven baselines. They not only achieve higher predictive accuracy but also strictly adhere to thermodynamic constraints, exhibiting near-zero Physical Violation Rates (PVR) and Maximal Overshoot (MO). Furthermore, we identify the cKAN architecture as highly resilient to severe sensor noise.

\end{itemize}

\section{Methodology}\label{sec2}

\begin{figure}[H]
    \centering
    \includegraphics[width=0.9\textwidth]{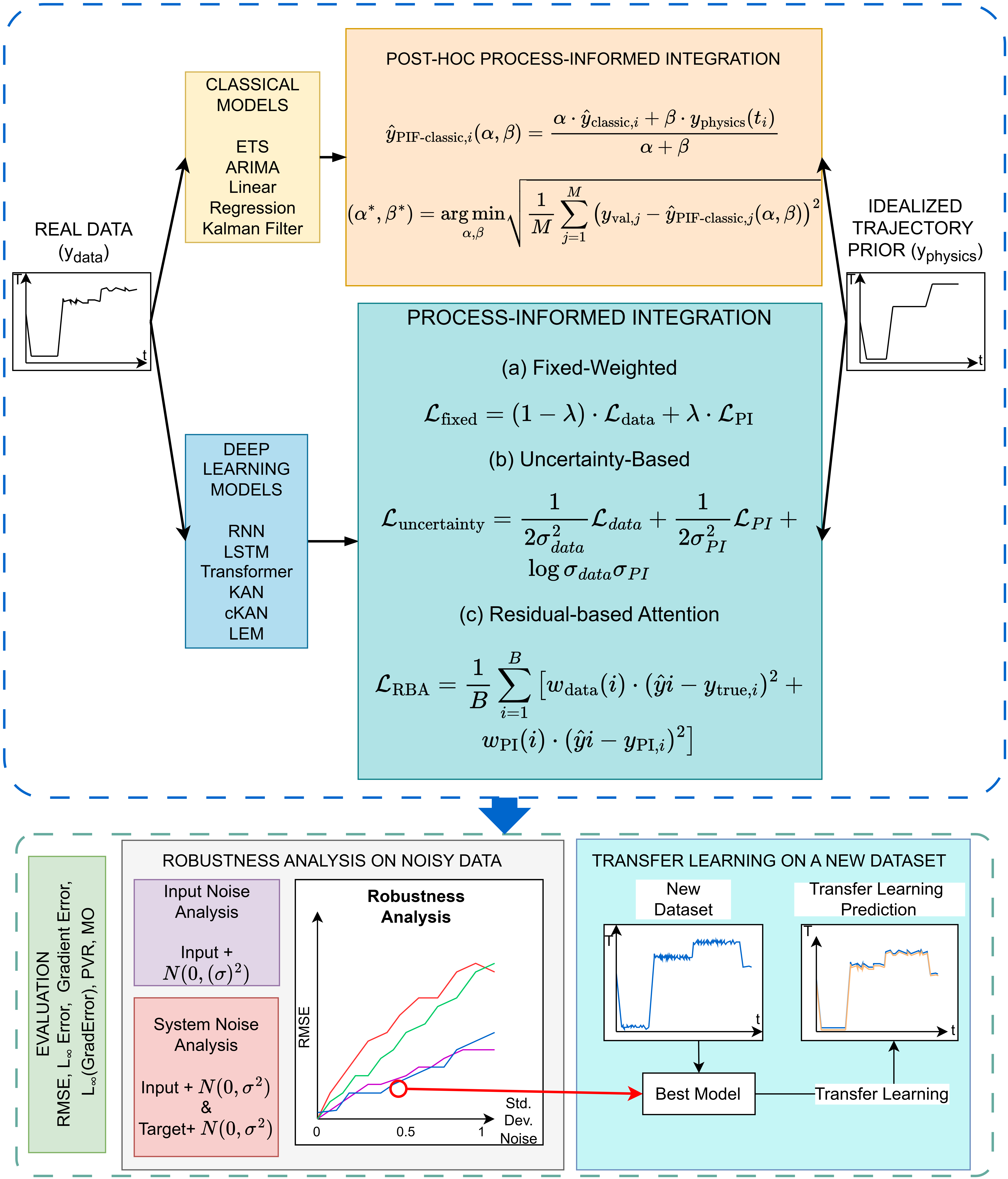}
    \caption{Overview of the proposed  Process-Informed Forecasting (PIF) methodology. (Upper part) Real sensor data $(y_\text{data})$ are used to train classical and deep learning models. A Process-Informed (PI) prior $(y_\text{PI})$, derived from the manufacturing recipe, is then incorporated to create PIF models. Classical models undergo post-hoc integration, while deep learning models are guided by three distinct loss function formulations: (a) Fixed-Weighted, (b) Uncertainty-Based, and (c) Residual-Based Attention (RBA). 
    (Lower Part) The performance of all models, classical, data-driven and PIF models, is assessed through standard metrics of accuracy $(RMSE$, $ L_\infty Error)$ and physical plausibility ($Gradient Error$, $L_\infty(GradErr)$, Physical Violation Rate (PVR \%) and Maximum Overshoot (MO)). Then, a robustness analysis is performed under two distinct noise injection scenarios: input-only and system-wide. Finally, a transfer learning task evaluates the generalizability of the best-performing model by applying it to a new, unseen dataset.
    }
    \label{fig:methodology}
\end{figure}

The methodological approach of this study, depicted in Figure~\ref{fig:methodology}, is designed to predict temperature dynamics in the pharmaceutical lyophilization process.
The framework is implemented in PyTorch and executed on an NVIDIA GPU, leveraging CUDA for computational acceleration. The dataset originates from a real-world pharmaceutical freeze-drying process, with the shelf inlet temperature as the primary target variable. The raw temperature data is sampled at one measurement per minute (1/60~Hz). Both the primary and secondary datasets are collected from the same industrial freeze-dryer: the primary dataset consists of 5 complete production batches for a first pharmaceutical product and the secondary dataset consists of 5 batches for a second product, manufactured at different periods. The data is partitioned chronologically into training (60\%), validation (20\%), and test (20\%) subsets to preserve temporal order and prevent information leakage. All features are normalized to a $[-1, 1]$ range.
 
We formulate a piecewise linear function that approximates the ideal temperature evolution during freeze-drying to serve as the Process-Informed (PI) prior. The PI prior is designed to balance physical realism with practical applicability. While a first-principles model based on thermodynamic equations would offer the highest fidelity, such models are often complex in real-world manufacturing settings. They require numerous product-specific parameters, e.g., vial heat transfer coefficients, product resistance to mass flow, and eutectic temperatures, that vary between products and are not available in historical process data. Therefore, we formulate an effective prior based on the known operational setpoints of the manufacturing recipe. This recipe-informed approach captures the dominant, low-frequency thermal dynamics defined by the process recipe, which provides the main temperature variation. Moreover, since lyophilization is a slow, multi-day process with gradual, controlled phase transitions, linear approximations of the ramps and holds are a reasonable and effective simplification. Our piecewise linear function models the idealized temperature trajectory as a series of ramps and holds corresponding to the freezing, primary drying and secondary drying phases.
 
\begin{equation}
y(t) = \begin{cases}
    y_0 + \frac{y_1 - y_0}{t_1 - t_0}(t - t_0), & \text{for } 0 \leq t < t_1 \quad\text{(freezing)} \\
    y_1, & \text{for } t_1 \leq t < t_2 \quad\text{(freezing)}\\
    y_1 + \frac{y_2 - y_1}{t_3 - t_2}(t - t_2), & \text{for } t_2 \leq t < t_3 \quad\text{(primary drying)} \\
    y_2, & \text{for } t_3 \leq t < t_4 \quad\text{(primary drying)}\\
    y_2 + \frac{y_3 - y_2}{t_5 - t_4}(t - t_4), & \text{for } t_4 \leq t < t_5 \quad\text{(secondary drying)}\\
    y_3, & \text{for } t_5 \leq t < t_6 \quad\text{(secondary drying)}
\end{cases}
\label{eq:piecewise_linear_function}
\end{equation}

\begin{table}[H]
\centering
\caption{Mapping of the time and temperature parameters in 
Equation~\eqref{eq:piecewise_linear_function} to the corresponding 
sub-phases of the lyophilization cycle.}
\label{tab:eq1_mapping}
% \tiny
\resizebox{\textwidth}{!}
{\begin{tabular}{llll}
\hline
\textbf{Parameter} & \textbf{Type} & \textbf{Physical meaning} & \textbf{Phase} \\
\hline
$t_0$ & Time & Start of lyophilization cycle        & Freezing (ramp) \\
$t_1$ & Time & End of freezing ramp                 & Freezing (hold) \\
$t_2$ & Time & Start of primary drying ramp         & Primary drying (ramp) \\
$t_3$ & Time & End of primary drying ramp            & Primary drying (hold) \\
$t_4$ & Time & Start of secondary drying ramp       & Secondary drying (ramp) \\
$t_5$ & Time & End of secondary drying ramp          & Secondary drying (hold) \\
$t_6$ & Time & End of lyophilization cycle           & Secondary drying (hold) \\
\hline
$y_0$ & Temperature & Initial shelf temperature setpoint   & Freezing (ramp) \\
$y_1$ & Temperature & Freezing hold setpoint               & Freezing (hold) \\
$y_2$ & Temperature & Primary drying setpoint              & Primary drying (hold) \\
$y_3$ & Temperature & Secondary drying setpoint            & Secondary drying (hold) \\
\hline
\end{tabular}}
\end{table}
 
Here, $y_0,y_1,y_2,\text{ and }y_3$ represent the temperature setpoints at the start of each main process phase, while $t_0,t_6$ correspond to the start and end times of the lyophilization cycle. The intermediate time values $t_1,t_2,...,t_5$ define the transition boundaries between the sub-phases, including the ramp-up, hold, and transition periods.
These parameters are directly extracted from the Production Recipe programmed into the lyophilizer's Programmable Logic Controller (PLC). Since every pharmaceutical product requires a validated recipe before production begins, these parameters are always available in industrial settings, making our Process-Informed prior highly practical and immediately deployable.

\begin{figure}[H]
    \centering
    {\includegraphics[width=\textwidth]{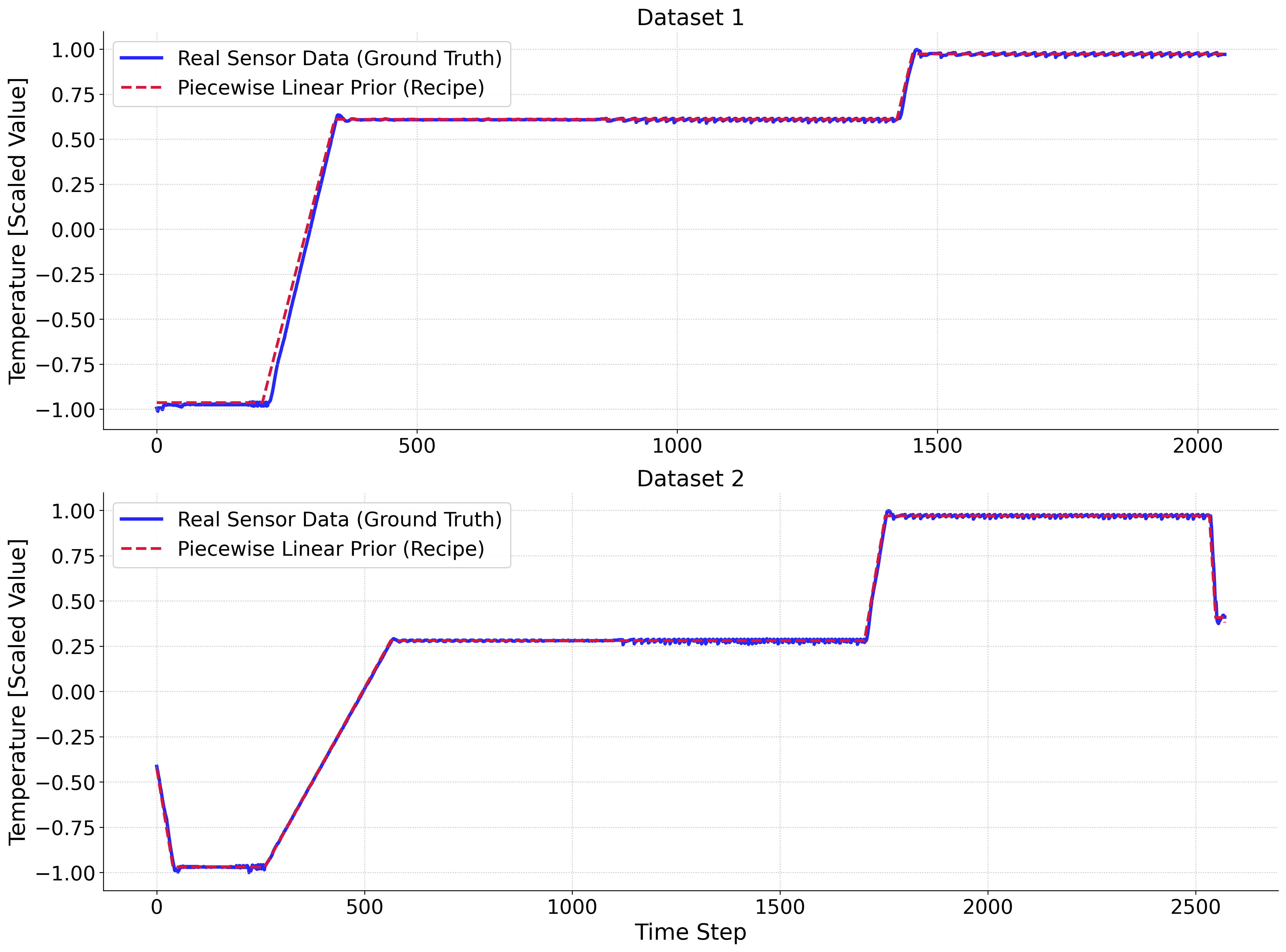}}
    \caption{Comparison between the real thermal dynamics (ground truth sensor data) and the idealized piecewise linear prior derived from the PLC production recipe. Dataset 1 exhibits pronounced thermal inertia during the primary ramp-up, which the PIF model is tasked with learning as a residual. Dataset 2 demonstrates a different thermal profile with a distinct ramp-up slope and an extended secondary drying phase. The close alignment between the recipe prior and the sensor data in both cases confirms that the piecewise linear formulation provides a robust macro-structural guide, regardless of the specific product-dependent thermal dynamics.}
    \label{fig:prior_vs_real}
\end{figure}

While we acknowledge that the underlying sublimation and desorption thermodynamics exhibit inherently non-linear behavior, the piecewise linear prior is not intended as a first-principles thermodynamic model but rather as a macro-structural control anchor. The lyophilizer's PID controllers actively force the system to track these linear setpoints, making the linear approximation an accurate descriptor of the controlled macro-dynamics rather than of the raw physics, see Figure~\ref{fig:prior_vs_real}. As the plot clearly shows, the real thermal dynamics closely follow the linear prior. While the thermal inertia is prominent in Dataset 1 due to the steeper ramp-up and higher thermal load, the prior remains a highly accurate structural anchor for Dataset 2. Our methodology relies on the neural network to learn this specific non-linear residual, while the linear prior provides the macro-structural guidance. This consistency across different products validates the generalizability of our process-informed approach.
 
This formulation enables the generation of a continuous, piecewise-linear target trajectory that captures the dominant thermal structure of the lyophilization process and accounts for practical operating constraints. 
 
Sequential input vectors for deep learning models are constructed using a lookback window of 300 time steps (i.e., the past 300~minutes of sensor data). All deep learning models perform one-step-ahead direct forecasting: given the past 300 observations, the model predicts the temperature at the next time step $t+1$.
 
We explore different loss function formulations. In classical models, we define the PIF as a convex combination of the model output $\hat{y}_{\text{classic}, i}$ and the process-informed prior $y(t_i)$. This is controlled by two non-negative weights, $\alpha$ and $\beta$, which determine the importance of the data-driven and process-informed components:
\begin{equation}
\hat{y}_{\text{PIF\_classic}, i} (\alpha, \beta) = \frac{\alpha \cdot \hat{y}_{\text{classic}, i} + \beta \cdot y(t_i)}{\alpha + \beta}
\end{equation}
Here, $\alpha$ is the trust in the classical model's prediction, while $\beta$ reflects the confidence of the process-informed prior. The normalization by $\alpha + \beta$ ensures that the result is a weighted average between the two extremes.
The optimal coefficients $\alpha$ and $\beta$ are selected by minimizing the Root Mean Square Error (RMSE) on a validation set:
\begin{equation}
(\alpha^*, \beta^*) = \underset{\alpha, \beta}{\arg\min} \sqrt{ \frac{1}{M} \sum_{j=1}^{M} \left( y_{\text{val}, j} - \hat{y}_{\text{PIF\_classic}, j}(\alpha, \beta) \right)^2 }
\end{equation}
 
For deep learning models, we investigate three distinct loss function formulations, progressing from a static weighting to more sophisticated dynamic schemes. First, we implement a fixed-weight loss function defined as
\begin{equation}
\mathcal{L}_{\text{fixed}} = (1 - \lambda) \cdot \mathcal{L}_{\text{data}} + \lambda \cdot \mathcal{L}_{\text{PI}}
\end{equation}

Here, $\mathcal{L}_{data}$ is the standard supervised loss defined as the Mean Squared Error (MSE) between the vector of model predictions $\hat{\mathbf{y}} \in \mathbb{R}^N$ and the vector of ground-truth sensor readings $\mathbf{y}_{\text{true}} \in \mathbb{R}^N$. The process-informed loss $\mathcal{L}_{\text{PI}}$ measures the MSE 
between $\hat{\mathbf{y}}$ and the vector of prior values $\mathbf{y}_{\text{PI}} \in \mathbb{R}^N$. For a batch of size $N$: 
\begin{align}
% \mathcal{L}_{\text{data}} &= \frac{1}{N} \sum_{i=1}^{N} (\hat{y}_i - y_{\text{true}, i})^2 &
% \mathcal{L}_{\text{PI}} &= \frac{1}{N} \sum_{i=1}^{N} (\hat{y}_i - y_{\text{PI}}(t_i))^2
\mathcal{L}_{\text{data}} = \frac{1}{N} \lVert \hat{\mathbf{y}} - \mathbf{y}_{\text{true}} \rVert^2
\qquad
\mathcal{L}_{\text{PI}} = \frac{1}{N} \lVert \hat{\mathbf{y}} - \mathbf{y}_{\text{PI}} \rVert^2
\end{align}
 
The hyperparameter $\lambda$ is determined empirically and establishes a static trade-off between data fidelity and process-informed consistency throughout the training process. Here, $\lambda$ is systematically optimized using the Optuna framework for automated hyperparameter search \cite{optuna_2019}. The search is conducted over a continuous range of $[0.01, 0.99]$ with a computational budget of 15 trials. The test set is strictly isolated throughout this phase to prevent data leakage. The optimal value for $\lambda$ is selected by minimizing the Mean Squared Error (MSE) exclusively on the validation data. The test set is reserved entirely and solely for the final evaluation of the trained models. This approach ensures that the chosen value of $\lambda$ is optimal with respect to the validation data rather than selected through manual trial and error.
 
Second, we employ a homoscedastic uncertainty-weight loss, which dynamically balances the data-driven and process-informed terms. Inspired by Kendall et al. (2018) \cite{Kendall_2018}, this approach learns the optimal global balance by interpreting the loss weights as learnable uncertainty parameters:
\begin{equation}
\mathcal{L}_{\text{uncertainty}} = \frac{1}{2\sigma_{data}^2} \mathcal{L}_{data} + \frac{1}{2\sigma_{PI}^2} \mathcal{L}_{PI} + \log\sigma_{data}\sigma_{PI}
\end{equation}
where $\sigma_{data}$ and $\sigma_{PI}$ represent the learnable homoscedastic uncertainty parameters for the data-driven and process-informed components, respectively. The final $\log$ term acts as a regularizer that penalizes the model from increasing uncertainty, preventing it from ignoring the tasks entirely.
 
Third, we explore a Residual-Based Attention (RBA) scheme, a gradient-less and computationally efficient method for assigning dynamic, local weights to individual points in the loss calculation \cite{anagnostopoulos2024}. The per-sample attention weight vectors $\boldsymbol{\lambda}_{\text{data}}, \boldsymbol{\lambda}_{\text{PI}} \in \mathbb{R}^B$ 
are updated at each training step $k$ as:

% \begin{align}
% \lambda_{\text{data}, i}^{k+1} &\leftarrow (1 - \eta) \lambda_{\text{data}, i}^{k} + \eta \cdot | \hat{y}_i^k - y_{\text{true},i}^k | &
% \lambda_{\text{PI}, i}^{k+1} &\leftarrow (1 - \eta) \lambda_{\text{PI}, i}^{k} + \eta \cdot | \hat{y}_i^k - y_{\text{PI},i}^k |
% \end{align}

% \begin{align}
% \lambda_{\text{data}, i}^{k+1} &\leftarrow (1 - \eta)\, \lambda_{\text{data}, i}^{k} 
% + \eta \cdot \lvert \hat{y}_i^k - y_{\text{true},i}^k \rvert, & 
% \lambda_{\text{PI}, i}^{k+1} &\leftarrow (1 - \eta)\, \lambda_{\text{PI}, i}^{k} 
% + \eta \cdot \lvert \hat{y}_i^k - y_{\text{PI},i}^k \rvert,
% \end{align}

\begin{equation}
\begin{split}
\lambda_{\text{data}, i}^{k+1} &\leftarrow (1 - \eta)\, \lambda_{\text{data}, i}^{k} 
+ \eta \cdot \lvert \hat{y}_i^k - y_{\text{true},i}^k \rvert, \\
\lambda_{\text{PI}, i}^{k+1} &\leftarrow (1 - \eta)\, \lambda_{\text{PI}, i}^{k} 
+ \eta \cdot \lvert \hat{y}_i^k - y_{\text{PI},i}^k \rvert,
\end{split}
\end{equation}

where $\eta \in (0,1)$ is a scalar smoothing factor (see Section \ref{sec3} for the value used in our experiments).

The loss function for a batch of size $B$, with normalized weight vectors 
$\mathbf{w}_{\text{data}}, \mathbf{w}_{\text{PI}} \in \mathbb{R}^B$, is:
% \begin{equation}
% \mathcal{L}_{\text{RBA}} = \frac{1}{B} \sum_{i=1}^{B} \left[ w_{\text{data}}(i) \cdot (\hat{y}_i - y_{\text{true},i})^2 + w_{\text{PI}}(i) \cdot (\hat{y}_i - y_{\text{PI},i})^2 \right]
% \end{equation}

\begin{equation}
\mathcal{L}_{\text{RBA}} = \frac{1}{B} \sum_{i=1}^{B} 
\left[ w_{\text{data}}(i) \cdot (\hat{y}_i - y_{\text{true},i})^2 
+ w_{\text{PI}}(i) \cdot (\hat{y}_i - y_{\text{PI},i})^2 \right]
\end{equation}

where the normalized weights are given by:

\begin{equation}
\begin{split}
    w_{\text{data}}(i) = \frac{\lambda_{\text{data}}(i)}{\lambda_{\text{data}}(i) + \lambda_{\text{PI}}(i)}, 
    \\
    w_{\text{PI}}(i) = \frac{\lambda_{\text{PI}}(i)}{\lambda_{\text{data}}(i) + \lambda_{\text{PI}}(i)}.
\end{split}
\end{equation}

In our experiments, the smoothing factor is set to $\eta=0.01$, yielding a decay factor of $(1-\eta)=0.999$. This value provides a slow, stable exponential moving average of the per-sample residuals, preventing oscillations in the attention weights while remaining responsive to persistent high-residual regions across training steps. Both weight vectors $\boldsymbol{\lambda}_{\text{data}}$ and $\boldsymbol{\lambda}_{\text{PI}}$ are initialized to $0.5$, ensuring an equal, unbiased weighting of the data-driven and process-informed loss components at the start of training.
 
The fixed-weight method provides a strong upper bound on accuracy but is computationally expensive, requiring a full hyperparameter search (15 Optuna trials $\times$ max 50 epochs each). It is therefore best suited for stable, well-characterized, and highly regulated processes where the one-time training cost is not a bottleneck. The Uncertainty-based and RBA methods require only a single training run (1 trial $\times$ max 50 epochs), offering near-optimal performance at a fraction of the cost. They are the recommended choice for scalable deployments, dynamic environments, or rapid adaptation to new manufacturing recipes via transfer learning.
 
We also analyze classical time-series forecasting models, including ARIMA, ETS, Linear Regression, and Kalman Filters. These models are effective for linear, seasonal, or trend-dominated behavior, which is common in many industrial processes. Second, the evaluated deep learning models are standard Recurrent Neural Networks (RNNs), LSTM networks, and attention-based Transformers, as well as recent approaches such as KAN, cKAN, and LEM. Each of these methods integrates process-informed constraints into the learning process, addressing the limitations of pure data-driven approaches. All deep learning models are trained and tuned to have approximately the same number of parameters to ensure a fair comparison. 

We evaluate models across predictive accuracy and physical plausibility over a test set of size $M$. Predictive accuracy is quantified through the Root Mean Square Error (RMSE):
\begin{equation}
\text{RMSE} = \sqrt{ \frac{1}{M} \sum_{j=1}^{M} (\hat{y}_j - y_{\text{true},j})^2 }
\end{equation}
To capture worst-case performance, we also compute the L-infinity error ($L_\infty Error$), defined as the maximum absolute prediction error across the test set:
\begin{equation}
L_\infty Error = \max_{j} |\hat{y}_j - y_{\text{true}, j}|
\end{equation}
Moreover, the physical plausibility of the forecast is assessed via the Gradient Error:
\begin{equation}
\text{GradientError} = \frac{1}{M} \sum_{j=1}^{M} |\nabla\hat{y}_j - \nabla y_{\text{true}, j}|
\end{equation}
and its worst-case counterpart $L_\infty(\text{GradError}) = \max_{j} |\nabla\hat{y}_j - \nabla y_{\text{true}, j}|$.
 
The Physical Violation Rate (PVR) measures phase-specific monotonicity \cite{kotary2021end}. During constant-temperature hold phases, the theoretical gradient of the setpoint is zero. This metric calculates the percentage of time steps where the model predicts a significant variation during these steady-state phases:
\begin{equation}
PVR = \frac{1}{|H|} \sum_{j \in H} \mathbb{I}(|\nabla \hat{y}_j| > \epsilon) \times 100
\end{equation}
where $H = \{j \mid \nabla y_{PI,j} = 0\}$ represents the set of indices corresponding to the hold phases, $|H|$ is the total number of such points, $\mathbb{I}$ is the indicator function, and $\epsilon$ is a strict tolerance threshold (set to $0.01$ in our scaled domain). A low PVR indicates that the model genuinely understands the physical logic of steady-state stability.
 
Moreover, the Maximum Overshoot (MO) measures critical limit violations. In control theory and time-series forecasting, overshoot universally quantifies the maximum positive deviation of a system's response above its target setpoint \cite{ogata2010modern}. The MO is defined as:
\begin{equation}
MO = \max_{j} \left( \max(0, \hat{y}_j - y_{PI,j}) \right)
\end{equation}
By evaluating both PVR and MO, we ensure that the forecasting models not only minimize point-wise errors but also strictly respect the operational safety boundaries of the manufacturing process.
 
To assess robustness, we perform a noise perturbation analysis. In real-world manufacturing processes, sensor readings are rarely perfect due to probe drift, wiring picking up interference, and even ground-truth measurements can contain errors. A forecasting model that performs well on clean data may fail when deployed in such a noisy environment, compromising product quality and safety. Gaussian noise with zero mean and increasing standard deviation up to $\sigma=1$ is added to the test set after normalization, so that $\sigma$ is expressed relative to the $[-1,1]$ scaled domain. Consequently, $\sigma=1$ represents a perturbation equal to 50\% of the entire peak-to-peak temperature variation of the lyophilization cycle. To contextualize this in physical terms, the shelf temperature spans approximately 80 \,°C from freezing to secondary drying. Thus, $\sigma =1$ in the normalized domain corresponds to a noise standard deviation of approximately $\pm40$\,°C on the raw signal - a magnitude consistent with catastrophic sensor failure rather than standard operational noise \cite{kim2015}.
 
We adopt two complementary strategies to assess different facets of robustness. In the first approach, Gaussian noise with increasing standard deviation is added only to the model input, while the target output remains unchanged. This simulates a typical industrial scenario in which sensor readings are noisy, while the ground-truth measurements used for evaluation are clean. Models that rely on sequential input, e.g. RNNs, LSTMs, KANs, are directly affected and must learn to predict with the noise. Classical models, which do not condition on input sensor data but instead rely on temporal coordinates, are immune to this perturbation. Thus, their prediction remains unchanged even when noise increases. This behavior is not a fault, but rather a demonstration of their independence from input noise, which can be an advantage in high-noise settings. The main strength of this test is that it isolates the model's internal robustness to input noise. However, it may overstate the reliability of time-conditioned models by failing to account for imperfections in the measured target.
 
Instead, in the second approach, we add noise to both the inputs and the target values. This models a situation where the sensor reading used for both prediction and evaluation is uncertain. Sequential models will be influenced by both sources. For classical models, while predictions remain constant, the RMSE varies depending on the amount of noise added to the ground truth, since they are being compared to increasingly uncertain targets. This second approach is more realistic for real-world deployments where ground truth values are not perfect. Together, these two strategies provide a rich and complementary understanding of model robustness.
 
Moreover, we extend the analysis to a transfer learning scenario. In pharmaceutical manufacturing, a single piece of equipment is often used to process multiple products, each with a unique processing recipe. Thus, developing and validating a new forecasting model from scratch for each product is both data-intensive and cost-prohibitive. Transfer learning is a solution that enables a robustly pre-trained model to be quickly adapted to a new task with minimal data and computational effort. We tested this capability by selecting the top-performing architecture from the robustness analysis at $\sigma=0.6$, which is a realistic level of sensor noise \cite{kim2015}. This pre-trained model is then applied to a new, unseen lyophilization dataset to evaluate its ability to adapt and perform in a novel context. We evaluate three transfer learning strategies. The first, Zero-Shot, applies the pre-trained model directly to the new dataset without any adaptation. The second, Linear Probing, freezes the entire pre-trained feature extractor and retrains only the final linear output layer on the new data. The third, Training From Scratch, initializes the architecture randomly and trains it on the secondary dataset with the same adaptation budget, providing a fully adapted reference baseline. The adaptation set consists of 3 complete lyophilization cycles (60\% of the 5-batch secondary dataset).
 
To further validate the generalizability of our findings, we introduce a secondary dataset of lyophilized samples. This dataset originates from a different pharmaceutical product with a distinct thermal profile, including different temperature setpoints and a longer secondary drying phase. It is partitioned using the same 60/20/20 chronological split for training, validation, and testing. This secondary process serves as a challenging test case to assess whether the conclusions drawn from our primary analysis hold across different manufacturing contexts. The full evaluation metrics for all models on this secondary dataset are provided in~\ref{app3}, Table~\ref{tab:evaluation_metrics_grouped_rmse_2}.
 
All deep learning models are trained for a maximum of 50 epochs using the Adam optimizer with weight decay $10^{-5}$. An early stopping mechanism (patience 10~epochs) monitors the validation RMSE for all model families, terminates training when no improvement is observed, and restores the best-performing weights. For the pure data-driven, Fixed-weight, and Uncertainty-based models, a ReduceLROnPlateau scheduler with patience of 5 epochs and a reduction factor of 0.1 is additionally applied to facilitate convergence after a plateau. The RBA models employ a dedicated training loop with a fixed learning rate; in this case, the per-sample dynamic weighting mechanism---which updates the attention weights $\lambda_{\text{data},i}$ and $\lambda_{\text{PI},i}$ at every step based on current residuals---provides an implicit adaptation of the effective gradient signal, making an external scheduler less critical. Learning rates are tailored per architecture: $5\times10^{-4}$ for Transformer models and $1\times10^{-3}$ for all others. All deep learning models are evaluated across four complexity scales: approximately 15,000, 30,000, 45,000, and 115,000 parameters.

\section{Results and Discussion}\label{sec3}

\subsection{Classical Time-Series Models}\label{subsec2}

To establish a rigorous benchmark and isolate the value added by machine learning, we first evaluate a non-ML baseline: the 'Recipe Prior'. This baseline assesses the predictive accuracy of the idealized piecewise linear prior ($y_{PI}$) used directly as a forecast, without any algorithmic intervention. As shown in Table \ref{tab:classical_models}, the Recipe Prior achieves an RMSE of 0.0258. By definition, it perfectly adheres to the process constraints, yielding a Physical Violation Rate (PVR) of 0.0\% and a Max Overshoot of 0.000.

ARIMA, ETS, Kalman Filter, and Linear Regression are tested in both their standard and process-informed forms. Through a validation-based grid search, the optimal configuration for the PI version is $\alpha=0.7$ and $\beta=0.3$. This weighting minimizes RMSE on the validation set and is applied consistently across all PI variants. The performance of the classical forecasting models is summarized in Table \ref{tab:classical_models}. The results show a trade-off between predictive accuracy and physical plausibility. In terms of pure forecasting accuracy, the standard models achieve lower $RMSE$ and $ L_\infty$ Error. Instead, the PIF models (\_informed), which incorporate the process-informed prior, achieve lower Gradient Error and significantly outperform the Recipe Prior baseline on $L_\infty(GradError)$. This demonstrates that the true thermal dynamics deviate from the ideal recipe due to thermal inertia, and classical models can learn to correct this residual error.

However, when evaluating physical plausibility, standard classical models exhibit significant limitations. For instance, the standard ARIMA model yields a PVR of 3.13\%, indicating frequent thermodynamically impossible temperature variations during the steady-state hold phases. Instead, the PIF models (\_informed), which incorporate the process-informed prior, drastically reduce these violations (e.g., the PVR for ARIMA\_informed drops to 0.05\%, and the Kalman Filter\_informed achieves 0.00\%). Furthermore, PIF variants consistently minimize the Max Overshoot, ensuring safer predictions.

\begin{table}[H]
\centering
\caption{Comparison of metrics for classical time-series models and their PIF variants. The Recipe Prior is included as a non-ML baseline. Standard models outperform their process-informed counterparts in forecasting accuracy. Instead, PIF models achieve better alignment with the process dynamics, achieving lower Gradient Error, and strictly adhering to phase-specific monotonicity (near-zero PVR).} \label{tab:classical_models}
% \tiny
\resizebox{\textwidth}{!}{
\begin{tabular}{lcccccccc}
\hline
\textbf{Model} & \textbf{RMSE} & \textbf{$L_\infty$Error} & \textbf{GradientError} & \textbf{$L_\infty$(Grad)} & \textbf{PVR (\%)} & \textbf{MO} & \textbf{Time(s)} \\
\hline
Recipe Prior Baseline & 0.0259 & 0.1618 & 0.0016 & 0.0210 & 0.00 & 0.000 & N/A \\
\hline
\textbf{ARIMA Models} & & & & & & & \\
\hline
ARIMA & \textbf{0.0047} & \textbf{0.0309} & 0.0015 & 0.0180 & 3.13 & 0.028 & 2207.4 \\
ARIMA\_informed & 0.0073 & 0.0493 & \textbf{0.0013} & \textbf{0.0174} & \textbf{0.05} & \textbf{0.019} & \textbf{2196.3} \\
\hline
\textbf{ETS Models} & & & & & & & \\
\hline
ETS & \textbf{0.0047} & \textbf{0.0371} & 0.0014 & 0.0184 & 2.60 & 0.036 & \textbf{840.9} \\
ETS\_informed & 0.0080 & 0.0486 & \textbf{0.0013} & \textbf{0.0173} & \textbf{0.10} & \textbf{0.025} & 841.4 \\
\hline
\textbf{Linear Regression} & & & & & & & \\
\hline
LinearRegression & \textbf{0.0048} & \textbf{0.0362} & 0.0017 & 0.0279 & 2.70 & 0.033 & \textbf{0.109} \\
LinearRegression\_informed & 0.0083 & 0.0540 & \textbf{0.0015} & \textbf{0.0217} & \textbf{0.26} & \textbf{0.023} & 0.195 \\
\hline
\textbf{Kalman Filter} & & & & & & & \\
\hline
KalmanFilter & \textbf{0.0054} & \textbf{0.0312} & 0.0019 & 0.0173 & 0.79 & 0.041 & \textbf{0.335} \\
KalmanFilter\_informed & 0.0085 & 0.0466 & \textbf{0.0016} & \textbf{0.0169} & \textbf{0.00} & \textbf{0.028} & 0.339 \\
\hline
\end{tabular}}
\end{table}

\subsection{Deep Learning Models}\label{subsec3}
Our investigation involves benchmarking seven families of deep learning models and their PIF variants across three levels of complexity: approximately 15,000, 30,000, 45,000 and 115,000 parameters. The detailed results, presented in Table \ref{tab:dl_models_15k}, Table \ref{tab:dl_models_30k}, Table \ref{tab:dl_models_45k} and Table \ref{tab:dl_models_115k},  provide insights regarding the interaction between model capability, architecture, and PI integration. 

Almost all evaluated deep learning models significantly outperform the Recipe Prior baseline (RMSE 0.0258), often halving the error (e.g., cKAN achieves an RMSE of 0.0086 at the 30,000 parameter scale). This proves that the neural networks learn to model the complex nonlinear thermal inertia residuals, actively using the prior as a macro-structural guide rather than forcing predictions to be linear. First, as the model size increases, the process-informed integration becomes more accurate. At the 15,000 parameter scale, different data-driven models (e.g. cKAN, KAN, MLP) have the lowest RMSE. However, as the parameter count grows to 30,000, 45,000 and 115,000 parameters, PIF models outperform their data-driven counterparts in terms of accuracy. At 115,000 parameters, a PI variant is the top performer in almost every model family. Thus, the process-informed prior is a strong regularizer, preventing higher-capacity models from overfitting to noise in the training data and guiding them towards more generalizable solutions.

\begin{figure}[H]
    \centering
    \includegraphics[width=\textwidth]{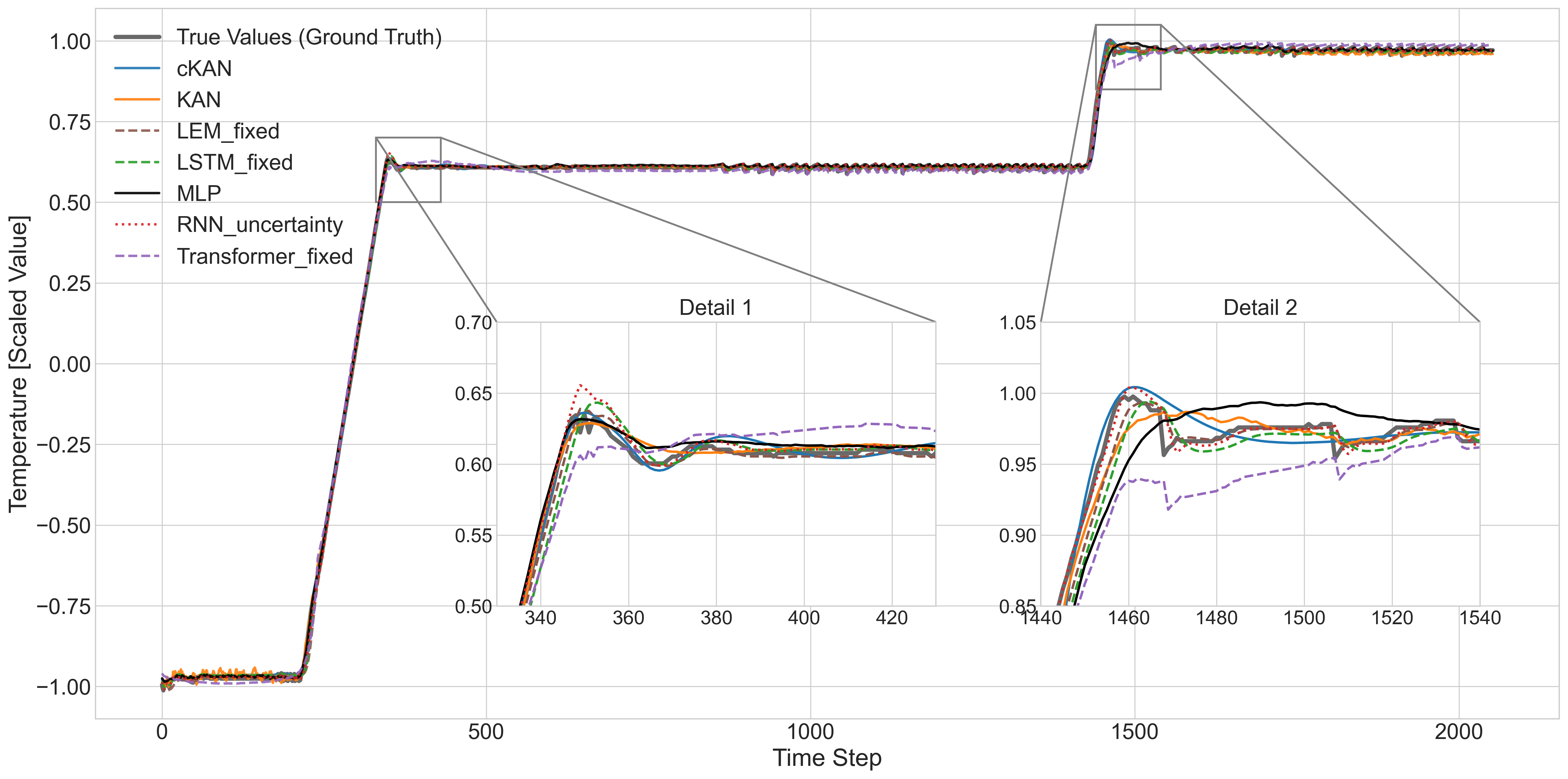}
    \caption{Comparative analysis of model predictions for the thermal dynamics of the lyophilization process. The figure shows the output of the best-performing variant from seven different model families ($\sim{30,000}$ parameters). These zoomed-in views highlight the effectiveness of architectures like cKAN, KAN and LEM in modeling non-linear, transient behavior. In contrast, the Transformer architecture struggles to capture the peak temperature, underscoring its limitations for this specific dynamic system.}
    \label{fig:30kpred}
\end{figure}

Furthermore, the evaluation of the physical metrics (PVR and Max Overshoot) highlights the strict adherence of PIF models to process constraints. Pure data-driven models often exhibit higher violation rates during steady-state hold phases. In contrast, PIF variants, particularly the fixed-weight MLP and cKAN, consistently achieve near-zero PVR and lower maximum overshoots, ensuring that the predictions remain thermodynamically safe. Notably, the standard Transformer model exhibits a 0.0\% PVR and low overshoot, but suffers from high RMSE and Gradient Error. This confirms that the global self-attention mechanism acts as a low-pass filter, oversmoothing predictions and failing to reach peak temperatures during rapid phase transitions, as shown in Figure \ref{fig:30kpred}. The Transformer's difficulty is rooted in the global self-attention mechanism: when computing the output for each time step, the model averages over the entire 300-step lookback window, inherently smoothing out the high-amplitude local gradients that characterize rapid phase transitions. Sequential architectures process time series step-by-step, assigning implicit causal weights to recent observations and capturing sudden changes more faithfully.

Moreover, while no single PIF strategy is universally superior across all architectures, the fixed-weight strategy (fixed) is often effective (e.g. for LEM and cKAN at 45,000 parameters). However, this strategy incurs high computational cost due to Optuna's extensive hyperparameter search. For complex models like KAN and LEM, this results in training times that are orders of magnitude longer than their data-driven counterparts. On the other hand, the uncertainty-based (uncertainty) and residual-based attention (RBA) methods are competitive and faster than the pure data-driven models. This makes them highly attractive candidates for real-world applications such as scalable deployments, dynamic environments, or rapid adaptation to new manufacturing recipes. The uncertainty-based models frequently rank among the best performers (e.g., for RNN and LSTM), offering a scalable approach to PIF methodology. 
While uncertainty-weighted models perform competitively on the first dataset, they exhibit reduced robustness to the second dataset compared to fixed-weight integration, see \ref{app3}. Figure \ref{fig:30kpred} provides a qualitative view of these performance differences for the top-performing models at the $\sim30,000$ parameter scale. While most models accurately track the stable phases, the zoom-in section reveals distinctions during the rapid temperature ramps. Architectures such as cKAN and KAN demonstrate strong ability to capture these sharp transients, closely following the ground truth. In contrast, the Transformer model fails to reach the peak temperature, visually confirming the limitations suggested by its higher Gradient Error metrics. 

\begin{table}[H]
\centering
\caption{Performance evaluation of deep learning models and their PIF models, with model complexity increased to approximately 30,000 parameters. The uncertainty-based PIF variant is the top performer for the RNN family; fixed-weighted models lead for LSTM and LEM; and the RBA strategy is the most accurate for the transformer. Uncertainty-based models and RBA methods confirm the computational efficiency trend, offering the fastest training times. Instead, the fixed-weighted strategy is accurate but often comes with a huge increase in training time (KAN, LEM).}\label{tab:dl_models_30k}%
% \tiny
\resizebox{\textwidth}{!}{
\begin{tabular}{lrrrrrrr}
\hline
\textbf{Model Family} & \textbf{RMSE} & \textbf{$L_\infty Error$} & \textbf{GradError} & \textbf{$L_\infty(Grad)$} & \textbf{PVR (\%)} & \textbf{MO} & \textbf{Time(s)} \\
\hline
Recipe Prior Baseline & 0.02586 & 0.16179 & 0.00162 & 0.02100 & 0.00 & 0.000 & N/A \\
\hline
\multicolumn{8}{l}{\textbf{RNN Models}} \\
\hline
RNN\_uncertainty & \textbf{0.0064} & \textbf{0.0333} & \textbf{0.0019} & 0.0163 & \textbf{0.4251} & 0.0454 & \textbf{7.2} \\
RNN\_RBA & 0.0085 & 0.0350 & 0.0019 & 0.0154 & 0.5313 & 0.0420 & 8.7 \\
RNN & 0.0097 & 0.0395 & 0.0019 & 0.0164 & 0.4782 & \textbf{0.0304} & 12.2 \\
RNN\_fixed & 0.0100 & 0.0378 & 0.0019 & \textbf{0.0150} & 0.4782 & 0.0319 & 155.8 \\
\hline
\multicolumn{8}{l}{\textbf{cKAN Models}} \\
\hline
cKAN & \textbf{0.0086} & 0.0705 & 0.0017 & \textbf{0.0145} & 0.2125 & 0.0308 & 9.9 \\
cPKAN\_fixed & 0.0094 & \textbf{0.0549} & \textbf{0.0016} & 0.0158 & \textbf{0.0531} & \textbf{0.0241} & 109.9 \\
cPKAN\_RBA & 0.0199 & 0.0886 & 0.0017 & 0.0176 & 0.3719 & 0.0431 & 9.6 \\
cPKAN\_uncertainty & 0.0205 & 0.1257 & 0.0018 & 0.0153 & 0.1063 & 0.0316 & \textbf{6.4} \\
\hline
\multicolumn{8}{l}{\textbf{KAN Models}} \\
\hline
KAN & \textbf{0.0093} & \textbf{0.0466} & 0.0020 & \textbf{0.0189} & 1.2221 & 0.0225 & 1835.9 \\
PKAN\_fixed & 0.0094 & 0.0480 & 0.0020 & 0.0189 & 1.2221 & 0.0224 & 29389.3 \\
PKAN\_uncertainty & 0.0147 & 0.0940 & \textbf{0.0019} & 0.0202 & \textbf{0.4782} & \textbf{0.0185} & \textbf{1808.7} \\
PKAN\_RBA & 0.0148 & 0.1061 & 0.0021 & 0.0209 & 1.0627 & 0.0232 & 1963.8 \\
\hline
\multicolumn{8}{l}{\textbf{LEM Models}} \\
\hline
LEM\_fixed & \textbf{0.0088} & 0.0790 & 0.0018 & \textbf{0.0153} & 0.4251 & 0.0287 & 29684.6 \\
LEM\_RBA & 0.0103 & \textbf{0.0610} & 0.0018 & 0.0158 & 0.3719 & 0.0262 & 2537.4 \\
LEM\_uncertainty & 0.0115 & 0.0650 & \textbf{0.0017} & 0.0162 & \textbf{0.2657} & 0.0274 & 1448.9 \\
LEM & 0.0119 & 0.0864 & 0.0018 & 0.0156 & 0.4251 & \textbf{0.0170} & \textbf{1447.2} \\
\hline
\multicolumn{8}{l}{\textbf{LSTM Models}} \\
\hline
LSTM\_fixed & \textbf{0.0097} & \textbf{0.0569} & \textbf{0.0019} & 0.0154 & \textbf{0.3188} & \textbf{0.0331} & 185.5 \\
LSTM & 0.0106 & 0.0735 & 0.0019 & 0.0156 & 0.3188 & 0.0340 & 15.6 \\
LSTM\_RBA & 0.0112 & 0.0598 & 0.0020 & 0.0150 & 0.4782 & 0.0385 & 13.7 \\
LSTM\_uncertainty & 0.0112 & 0.0622 & 0.0020 & \textbf{0.0153} & 0.3719 & 0.0354 & \textbf{10.0} \\
\hline
\multicolumn{8}{l}{\textbf{MLP Models}} \\
\hline
MLP & \textbf{0.0097} & \textbf{0.0660} & \textbf{0.0017} & \textbf{0.0181} & 0.1594 & 0.0215 & 10.3 \\
MLP\_uncertainty & 0.0151 & 0.0834 & 0.0018 & 0.0189 & \textbf{0.0000} & 0.0226 & 10.3 \\
MLP\_fixed & 0.0169 & 0.1210 & 0.0018 & 0.0192 & 0.0000 & \textbf{0.0191} & 102.4 \\
MLP\_RBA & 0.0177 & 0.0858 & 0.0018 & 0.0185 & 0.0000 & 0.0295 & \textbf{8.6} \\
\hline
\multicolumn{8}{l}{\textbf{Transformer Models}} \\
\hline
Transformer\_fixed & \textbf{0.0165} & \textbf{0.0659} & \textbf{0.0015} & 0.0258 & 0.0531 & 0.0212 & 647.7 \\
Transformer\_RBA & 0.0172 & 0.0977 & 0.0016 & 0.0397 & \textbf{0.0000} & 0.0290 & 66.8 \\
Transformer\_uncertainty & 0.0199 & 0.1205 & 0.0015 & \textbf{0.0171} & 0.1063 & 0.0303 & \textbf{31.6} \\
Transformer & 0.0206 & 0.1190 & 0.0015 & 0.0285 & 0.2125 & \textbf{0.0145} & 46.8 \\
\hline
\end{tabular}}
\end{table}

The results in Table~\ref{tab:summary_ranking} highlight three distinct performance profiles. cKAN and MLP achieve the most balanced profiles: cKAN leads in accuracy and physical consistency (lowest GradError and PVR) while  RNN\_uncertainty achieves the lowest RMSE (0.0064). MLP\_uncertainty, MLP\_fixed, MLP\_RBA, and Transformer\_RBA achieve a PVR of 0.00\%. LSTM\_fixed and LEM\_fixed offer competitive accuracy, but the latter's training time limits its practical scalability. KAN shows the highest GradError and PVR among the top-performing families, suggesting relatively lower adherence to steady-state constraints. The Transformer\_fixed presents a distinctive trade-off: it achieves the best GradError (0.00147) and second-lowest PVR, reflecting the smoothing effect of global self-attention, but at the expense of accuracy (RMSE 0.0165, nearly double that of cKAN) and high training costs. Figure~\ref{fig:30kpred} provides a qualitative confirmation of these findings.

\begin{table}[H]
\centering
\caption{Summary of the best-performing variant per model family at $\sim$30,000 parameters. RMSE and $L_\infty$Error measure predictive accuracy; Gradient Error and $L_\infty$(Grad) assess the physical plausibility of the predicted dynamics; PVR and MO quantify adherence to process safety constraints; Time reports the training time of the best variant. Models are ordered by overall performance profile.}\label{tab:summary_ranking}
% \tiny
\resizebox{\textwidth}{!}{
\begin{tabular}{llcccccccc}
\hline
\textbf{Best Variant} & \textbf{RMSE} & \textbf{$L_\infty$Error} & \textbf{GradError} & \textbf{$L_\infty$(Grad)} & \textbf{PVR (\%)} & \textbf{MO} & \textbf{Time (s)} \\
\hline
cKAN               & 0.0086 & 0.0705 & 0.00167 & \textbf{0.0145} & 0.21 & 0.031 & 9.9     \\
MLP                & 0.0097 & 0.0660 & 0.00172 & 0.0181 & \textbf{0.00} & 0.022 & 10.3    \\
RNN\_uncertainty   & \textbf{0.0064} & \textbf{0.0333} & 0.00190 & 0.0163 & 0.43 & 0.045 & \textbf{7.2}     \\
LSTM\_fixed        & 0.0097 & 0.0569 & 0.00194 & 0.0154 & 0.32 & 0.033 & 185.5   \\
LEM\_fixed         & 0.0088 & 0.0790 & 0.00177 & 0.0153 & 0.43 & 0.029 & 29684.6 \\
KAN                & 0.0093 & 0.0466 & 0.00204 & 0.0189 & 1.22 & 0.023 & 1835.9  \\
Transformer\_fixed & 0.0165 & 0.0659 & \textbf{0.00147} & 0.0258 & 0.05 & \textbf{0.021} & 647.7  \\
\hline
\end{tabular}}
\end{table}

\begin{table}[H]
\centering
\caption{Performance metrics for high-capacity models (approx. 115,000 parameters). The results show that increasing model capacity does not guarantee better performance. At this higher capacity, the regularizing effect of adaptive PIF strategies becomes more visible, with the RNN\_uncertainty and LEM models emerging as top performers. The results also underscore the computational cost of the fixed-weight strategy for complex models like KAN.}\label{tab:dl_models_115k}%
% \tiny
\resizebox{\textwidth}{!}{
\begin{tabular}{lrrrrrrr}
\hline
\textbf{Model Family} & \textbf{RMSE} & \textbf{$L_\infty Error$} & \textbf{GradError} & \textbf{$L_\infty(Grad)$} & \textbf{PVR (\%)} & \textbf{MO} & \textbf{Time(s)} \\
\hline
Recipe Prior Baseline & 0.02586 & 0.16179 & 0.00162 & 0.02100 & 0.00 & 0.000 & N/A \\
\hline
\multicolumn{8}{l}{\textbf{RNN Models}} \\
\hline
RNN\_uncertainty & \textbf{0.0086} & 0.0584 & \textbf{0.0019} & 0.0164 & 0.3719 & 0.0396 & \textbf{13.1} \\
RNN\_RBA & 0.0159 & 0.0682 & 0.0019 & 0.0146 & 0.2657 & 0.0422 & 16.7 \\
RNN & 0.0095 & 0.0612 & 0.0019 & \textbf{0.0146} & \textbf{0.2657} & \textbf{0.0300} & 22.7 \\
RNN\_fixed & 0.0113 & \textbf{0.0499} & 0.0020 & 0.0147 & 0.2657 & 0.0404 & 249.1 \\
\hline
\multicolumn{8}{l}{\textbf{cKAN Models}} \\
\hline
cPKAN\_fixed & \textbf{0.0543} & \textbf{0.1003} & 0.0019 & 0.0212 & 0.1594 & 0.0979 & 72.6 \\
cPKAN\_RBA & 0.0211 & 0.1307 & \textbf{0.0018} & 0.0188 & 0.4251 & 0.0367 & 5.0 \\
cPKAN\_uncertainty & 0.0864 & 0.2909 & 0.0025 & \textbf{0.0187} & 0.5845 &  0.1308 & \textbf{2.5} \\
cKAN & 0.0980 & 0.4043 & 0.0020 & 0.0203 & \textbf{0.0000} & \textbf{0.0437} & 4.7 \\
\hline
\multicolumn{8}{l}{\textbf{KAN Models}} \\
\hline
KAN & \textbf{0.0089} & \textbf{0.0492} & 0.0019 & \textbf{0.0173} & 0.1063 & 0.0182 & \textbf{6318.7} \\
PKAN\_fixed & 0.0091 & 0.0550 & 0.0019 & 0.0174 & 0.1063 & 0.0179 & 104168.6 \\
PKAN\_uncertainty & 0.0141 & 0.0968 & \textbf{0.0017} & 0.0161 & \textbf{0.0000} & \textbf{0.0126} & 6422.6 \\
PKAN\_RBA & 0.0147 & 0.1065 & 0.0019 & 0.0168 & 0.1063 & 0.0157 & 7868.8 \\
\hline
\multicolumn{8}{l}{\textbf{LEM Models}} \\
\hline
LEM\_fixed & 0.0099 & 0.0887 & 0.0019 & \textbf{0.0150} & 0.4251 & \textbf{0.0278} & 42803.5 \\
LEM\_RBA & 0.0106 & \textbf{0.0584} & 0.0019 & 0.0157 & \textbf{0.2657} & 0.0367 & 3277.9 \\
LEM\_uncertainty & 0.0126 & 0.0620 & \textbf{0.0018} & 0.0164 & 0.3188 & 0.0368 & \textbf{2228.7} \\
LEM & \textbf{0.0096} & 0.0893 & 0.0019 & 0.0152 & 0.3719 & 0.0282 & 2279.2 \\
\hline
\multicolumn{8}{l}{\textbf{LSTM Models}} \\
\hline
LSTM\_fixed & \textbf{0.0126} & 0.0838 & \textbf{0.0021} & 0.1590 & \textbf{0.3188} & 0.0357 & 286.8 \\
LSTM & 0.0141 & 0.1281 & 0.0021 & \textbf{0.0154} & 0.3188 & \textbf{0.0356} & 23.1 \\
LSTM\_RBA & 0.0138 & \textbf{0.0766} & 0.0021 & 0.1630 & 0.3719 & 0.0492 & \textbf{12.9} \\
LSTM\_uncertainty & 0.0137 & 0.0952 & 0.0021 & 0.0166 & 0.3188 & 0.0356 & 21.9 \\
\hline
\multicolumn{8}{l}{\textbf{MLP Models}} \\
\hline
MLP & \textbf{0.0140} & \textbf{0.0833} & 0.0019 & 0.0184 & 0.1063 & 0.0337 & 13.5 \\
MLP\_uncertainty & 0.0144 & 0.0838 & \textbf{0.0017} & \textbf{0.0183} & \textbf{0.0000} & 0.0197 & \textbf{9.6} \\
MLP\_fixed & 0.0120 & 0.0883 & 0.0018 & 0.0189 & 0.0000 & 0.0225 & 120.7 \\
MLP\_RBA & 0.0192 & 0.1053 & 0.0018 & 0.1865 & 0.0000 & \textbf{0.0165} & 9.6 \\
\hline
\multicolumn{8}{l}{\textbf{Transformer Models}} \\
\hline
Transformer\_fixed & 0.0138 & \textbf{0.0704} & 0.0015 & 0.0348 & \textbf{0.0000} & 0.0159 & 1006.4 \\
Transformer\_RBA & 0.0223 & 0.1271 & 0.0015 & 0.0164 & 0.1594 & 0.0429 & 63.3 \\
Transformer\_uncertainty & \textbf{0.0126} & 0.0780 & 0.0016 & \textbf{0.0158} & 0.0000 & 0.0102 & 78.1 \\
Transformer & 0.0247 & 0.1299 & \textbf{0.0014} & 0.0437 & 0.4251 & \textbf{0.0082} & \textbf{35.2} \\
\hline
\end{tabular}}
\end{table}

\subsection{Robustness Analysis}\label{subsec4}

\begin{figure}[H]
    \centering
    \includegraphics[width=\textwidth]{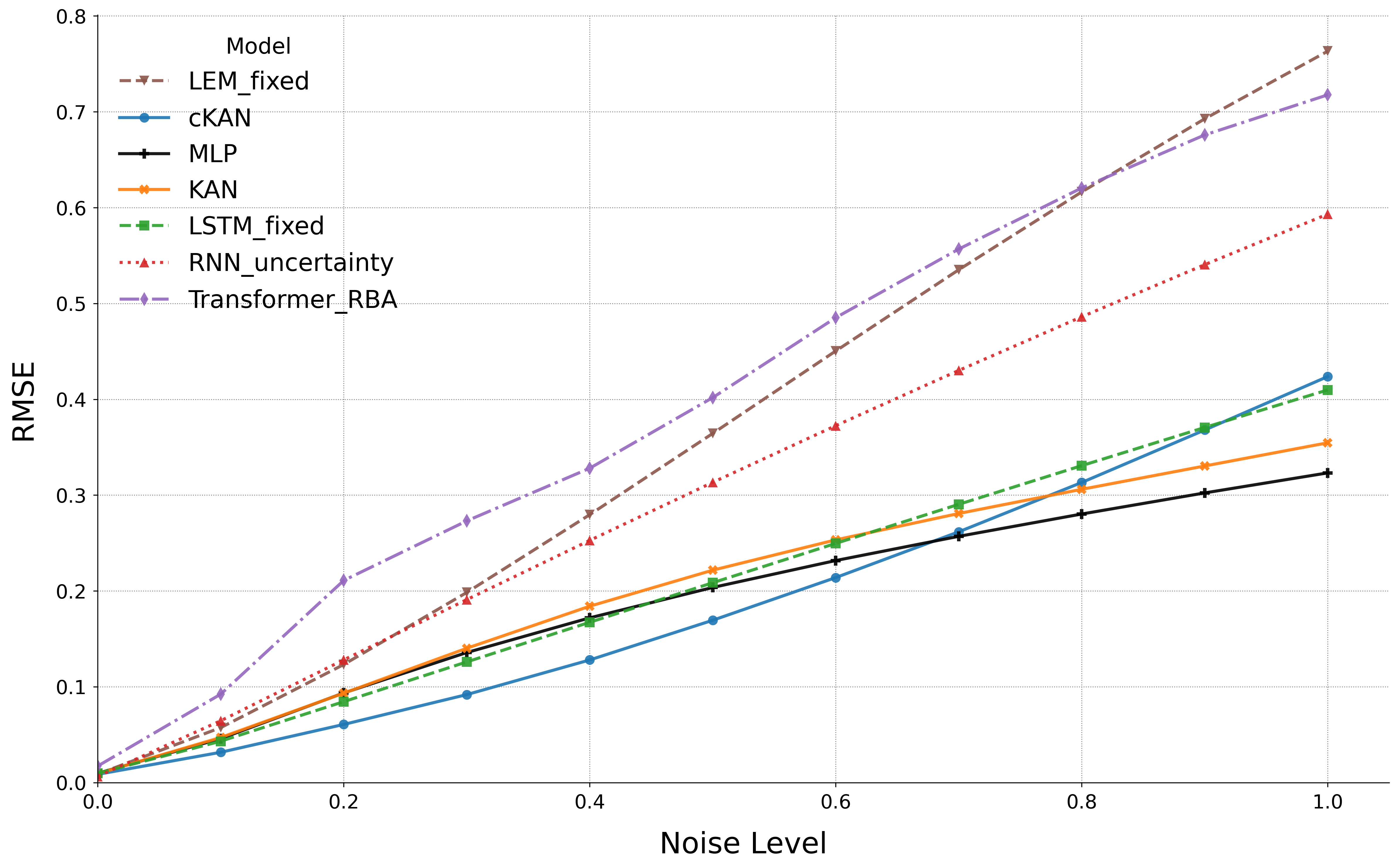}
    \caption{Robustness evaluation of models of approximately 30,000 parameters. This chart evaluates how the predictive accuracy of different models deteriorates as the input data becomes noisier. The results identify the cKAN model as the most resilient, maintaining the lowest error across the noise spectrum. The standard KAN and MLP demonstrate a strong ability to handle noise. Instead, architectures such as LEM\_fixed and Transformer\_RBA are more sensitive to data perturbations, with a rapid decline in accuracy as noise levels increase.}
    \label{fig:robust}
\end{figure}

We conducted a robustness analysis to assess the models' reliability under realistic operational conditions. We first evaluated the model's robustness to noisy sensor readings with only input noise. Figure \ref{fig:robust} shows the performance degradation of the best-performing model from each family at 30,000 parameters as a function of increasing input noise. The results identify the cKAN model as the most resilient, maintaining the lowest RMSE across the noisy spectrum. The noise robustness of cKAN relative to standard KAN can be attributed to the properties of its Chebyshev polynomial basis functions, which are globally defined and orthogonal, in contrast to the locally-supported B-splines of standard KAN. Combined with the $\tanh$ input normalization, this orthogonality acts as a natural low-pass filter, distributing the influence of any perturbation across the entire function approximation and suppressing high-frequency noise. Furthermore, we assessed physical plausibility under noise using the Physical Violation Rate (PVR) and Max Overshoot (MO) metrics. As shown in Table \ref{tab:noise_input_0600}, PIF models exhibit superior adherence to thermodynamic constraints compared to data-driven baselines, confirming that the prior acts as a structural anchor even under severe perturbation.
While KAN and MLP also demonstrate strong robustness, architectures such as LEM and Transformer with PI priors are more sensitive to input perturbations. Table \ref{tab:noise_input_0600} provides a detailed numerical breakdown at a plausible industrial environment level of noise $(\sigma=0.6)$ \cite{kim2015}, confirming that cKAN not only has a low average error (RMSE) but also excels at capturing the system's rate of change (Gradient Error). 

\begin{figure}[H]
    \centering
    \includegraphics[width=\textwidth]{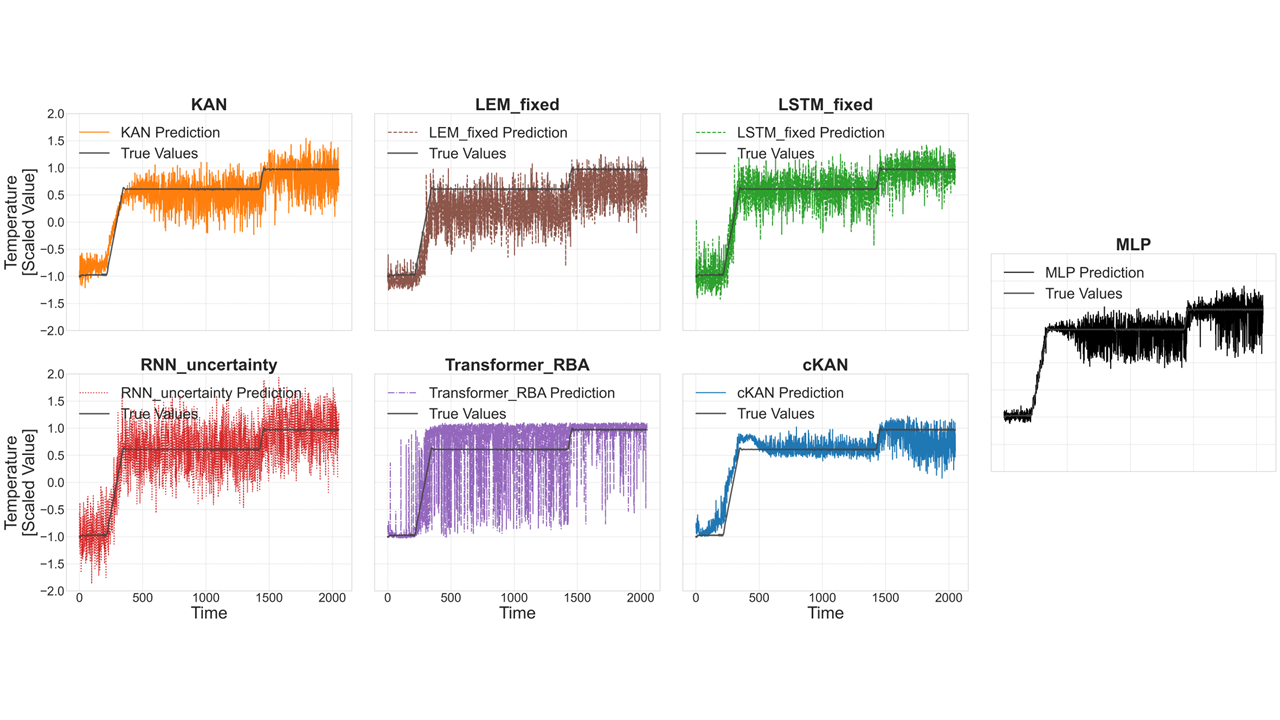}
    \caption{Model prediction comparison on noisy data (Noise $\sigma=0.6$). This figure shows the predictions of the best-performing models from each family ($\sim30,000$ parameters) on a test set where the input features are corrupted with Gaussian Noise ($\sigma=0.6$). The predictions are plotted against the original, noise-free ground truth to assess noise-filtering capability. The cKAN model proves to be robust, capable of ignoring fluctuations and predicting the true process state. In contrast, Transformer and RNN predictions indicate high variance, making them unsuitable for critical control and monitoring tasks.}
    \label{fig:best30krobust}
\end{figure}

To provide a qualitative illustration of the results, Figure \ref{fig:best30krobust} compares the models' predictions directly on a noisy input signal. The cKAN model effectively filters high-frequency noise, producing a smooth prediction that closely matches the noise-free ground truth. In contrast, the predictions from the RNN model become highly volatile, indicating that it is overfitting to the noise. To test if these findings are specific to our primary process, we have performed the same input noise analysis on the secondary lyophilization dataset. The results, detailed in \ref{app3} (Figure \ref{fig:robust_input_2dataset}, Table \ref{tab:noise_input_0600_2dataset}), reveal a consistent but slightly different pattern. While the same group of architectures (MLP, KAN-family) remains the most robust, the MLP model demonstrates the best overall resilience on this secondary process. It suggests that while certain architectures are inherently more robust, the optimal choice can be process-dependent. Figure \ref{fig:robustsubplot} shows a detailed breakdown for each model family. The classical models are immune to input noise, as their predictions are not conditioned on sensor inputs. For the deep learning families, the performance curves of the different variants are often clustered, confirming that the base architecture is the main factor of noise robustness. However, subtle differences emerge: the uncertainty-based PIF strategy demonstrates superior resilience for the cKAN (b) and Transformer (h) architectures, exhibiting a flatter degradation curve. In contrast, for the LEM (d) family, the pure data-driven model proves to be the most resilient. These findings indicate that while the base architecture dictates the sensitivity to noise, process-informed integration—specifically via uncertainty-based weighting—can act as a secondary stabilizer for complex, high-frequency-capable architectures.

\begin{figure}[H]
    \centering
    \includegraphics[width=\textwidth]{figure5.png}
    \caption{A detailed comparison of model robustness to input noise, with performance degradation (RMSE) analyzed separately for each model family. (a) classical time-series models (b) cKAN architectures; (c) KAN models; (d) LEM architectures; (e) LSTM models; (f) MLP architectures; (g) RNN models; (h) Transformer models. The noise level represents the standard deviation of Gaussian noise added to the input sensor features, while the target output remains unperturbed.}
    \label{fig:robustsubplot}
\end{figure}

\begin{table}[H]
\centering
\caption{A detailed numerical breakdown of model performance at the extreme 0.60 input noise level. The table evaluates predictive accuracy (RMSE), the ability to model system dynamics (Gradient Error), and adherence to thermodynamic constraints (PVR and Max Overshoot). While extreme input noise inevitably degrades physical consistency across all models (causing high PVR as models react to the noisy input), the results highlight distinct architectural strengths. The cKAN architecture achieves the lowest average error and provides the most accurate estimation of the system's rate of change. Conversely, the MLP demonstrates superior safety by maintaining the absolute lowest Max Overshoot (0.4316), preventing dangerous temperature spikes despite the severe sensor perturbation.}\label{tab:noise_input_0600}
\resizebox{\textwidth}{!}{
\begin{tabular}{lrrrrrr}
\hline
\multicolumn{7}{c}{\textbf{Noise Level: 0.60}} \\
\hline
\textbf{Model} & \textbf{RMSE} & \textbf{$L_\infty$Error} & \textbf{GradError} & \textbf{$L_\infty$(Grad)} & \textbf{PVR (\%)} & \textbf{MO} \\
\hline
cKAN & \textbf{0.2139} & 0.9029 & \textbf{0.0730} & \textbf{0.4965} & 88.58 & 0.7125 \\
MLP & 0.2317 & \textbf{0.8806} & 0.1120 & 0.6182 & 91.39 & \textbf{0.4316} \\
LSTM\_fixed & 0.2496 & 1.0438 & 0.1380 & 0.7165 & 95.38 & 1.0078 \\
KAN & 0.2533 & 1.0229 & 0.1219 & 0.6457 & 94.31 & 0.5722 \\
RNN\_uncertainty & 0.3725 & 1.1703 & 0.2101 & 0.8640 & 96.65 & 1.0977 \\
LEM\_fixed & 0.4504 & 1.4166 & 0.1574 & 0.7849 & 96.55 & 0.6901 \\
Transformer\_RBA & 0.4852 & 1.8779 & 0.1955 & 1.0371 & \textbf{83.42} & 1.8440 \\
\hline
\end{tabular}
}
\end{table}

\begin{figure}[H]
    \centering
    \includegraphics[width=\textwidth]{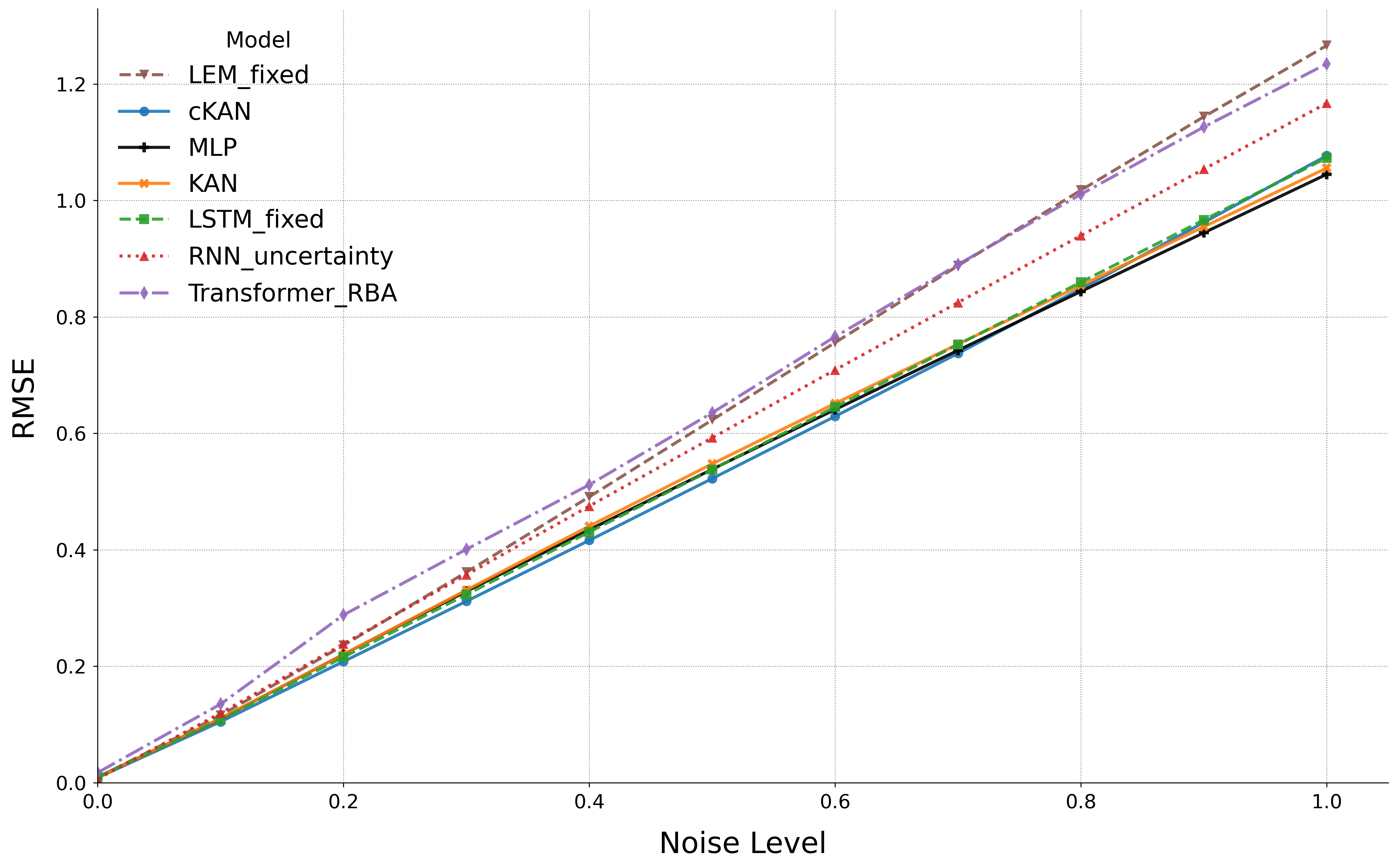}
    \caption{System-wide noise robustness evaluation of models of approximately 30,000 parameters. This chart illustrates how the performance (RMSE) of top models degrades when Gaussian noise is simultaneously added to both the input features (X) and the target values (y). The most robust models - MLP, cKAN, KAN - are clustered. Unlike the input-only noise test, where model performance diverged significantly, this scenario shows that when the evaluation target is also noisy, the differences in resilience between top architectures become less pronounced. }
    \label{fig:robust_global}
\end{figure}

Moreover, we have conducted a system-wide noise analysis on the primary dataset, adding noise to both the input and target values (Figure \ref{fig:robust_global}). This scenario shows that when the evaluation target is also uncertain, the performance differences between the top architectures become less pronounced, though the MLP, cKAN, and KAN models remain the most robust group.

\subsection{Transfer Learning for Process Adaptability}\label{subsec4}
For the final transfer learning phase, we have selected the best-performing robust model from our primary analysis, the pure cKAN with 30,000 parameters. This choice simulates a realistic workflow in which a model developed and validated for one process is then adapted to a new one. We have applied this pre-trained model to a new, unseen lyophilization dataset for a different pharmaceutical product. This new process shares the same underlying structure describable by the idealized trajectory prior (Equation \ref{eq:piecewise_linear_function}) but has a different set of setpoints and an additional drying phase. Thus, this new dataset makes the transfer learning process a realistic generalization challenge.
We evaluated three strategies to test the model's adaptability, see Table \ref{tab:transfer_learning}. The first strategy, Zero-Shot Baseline, uses a pre-trained model applied directly to the new dataset without retraining. Then, Linear Probing freezes the entire pre-trained feature extractor and trains only the final linear output layer on the new data. Finally, Training From Scratch trains a randomly initialized model on the new data with the same adaptation budget.
 
The Zero-Shot baseline yields the highest error (RMSE 0.1674), confirming that direct application without adaptation is insufficient. The Linear Probing strategy achieves highly competitive performance (RMSE=0.0142) by updating only the final output layer and achieves the lowest $L_\infty$ error (0.1247). This indicates that the frozen pre-trained layers encode generalizable physical representations that limit worst-case prediction deviations. Training From Scratch achieves the lowest RMSE (0.0116) and best Gradient Error (0.0027) by allowing all parameters to specialize to the new process and is the recommended strategy when adaptation data and compute are available. The Physical Violation Rate values reflect the wider temperature setpoint range of the secondary dataset, which is independently normalized to $[-1, 1]$ from a different physical temperature span and are therefore not directly comparable to the values reported for the primary dataset.
As evidenced by the results on the secondary dataset (Appendix~\ref{app3}, Table~\ref{tab:evaluation_metrics_grouped_rmse_2}), uncertainty-weighted models, while competitive on the source domain, consistently exhibit reduced robustness to domain shifts compared to fixed-weight integration. The uncertainty parameters $\sigma_{\text{data}}$ and $\sigma_{\text{PI}}$ learned on the source process encode a data-physics balance that may not transfer to a process with different thermal dynamics, setpoints, or phase durations. Thus, for multi-product deployments, the fixed-weight or RBA strategies should be preferred, as they demonstrate more stable cross-domain performance.

\begin{table}[H]
\centering
\caption{Performance comparison of transfer learning strategies for the pre-trained cKAN model (30,000 parameters) applied to a new, unseen lyophilization dataset. Three strategies are compared: Zero-Shot (no adaptation), Linear Probing (only the output layer retrained), and Training From Scratch (full retraining on the secondary data). Linear Probing achieves a 91.5\% reduction in RMSE relative to Zero-Shot, demonstrating that the pre-trained feature extractor encodes generalizable process representations. Training From Scratch achieves the lowest RMSE (0.0116) and Gradient Error by updating all parameters. Physical Violation Rate values reflect the wider temperature setpoint range of the secondary dataset (independently normalized to $[-1,1]$) and are not directly comparable to the primary-dataset MO.
}\label{tab:transfer_learning}
\resizebox{\textwidth}{!}{
\begin{tabular}{lrrrrrrr}
\hline
\textbf{Strategy} & \textbf{RMSE} & \textbf{$L_\infty$Error} & \textbf{GradError} & \textbf{$L_\infty$(Grad)} & \textbf{PVR (\%)} & \textbf{MO} & \textbf{Time(s)} \\
\hline
Zero-Shot (Baseline)  & 0.1674 & 0.6132 & 0.0032 & 0.0690 & 1.1590 & 0.6084 & \textbf{0.50} \\
Linear Probing        & 0.0142 & 0.1247 & 0.0029 & 0.0563 & 0.2782 & 0.2201 & 20.24 \\
Training From Scratch & \textbf{0.0116} & \textbf{0.1201} & \textbf{0.0027} & \textbf{0.0494} & \textbf{0.1854} & \textbf{0.2154} & 7.83 \\
\hline
\end{tabular}
}
\end{table}

\section{Summary}\label{sec13}

In this work, we address the challenge of developing reliable time-series forecasting models for complex thermal dynamics with high-frequency features in pharmaceutical manufacturing. We introduce a Process-Informed Forecasting (PIF) methodology to enhance predictive accuracy and physical plausibility of forecasting models. This effect is highlighted in higher-capacity deep learning models, where the PI prior acts as a regularizer, preventing overfitting and guiding the models toward more generalizable solutions. Our key findings are threefold. First, process-informed models outperform their purely data-driven counterparts, especially as model complexity increases, and PIF variants consistently exhibit superior physical plausibility as quantified by PVR and Maximum Overshoot. Second, our robustness analysis reveals that the intrinsic architecture of a model, such as cKAN's orthogonal Chebyshev basis, is a primary determinant of its resilience to sensor noise. Third, the transfer learning study confirms the practical utility of our approach, showing that a pre-trained cKAN model can act as a powerful feature extractor, enabling rapid adaptation to a new manufacturing process via linear probing with minimal computational effort. 
 
While the piecewise linear prior is highly effective under normal operating conditions, it assumes the lyophilizer's control system is functioning correctly. A key limitation of this static prior is its vulnerability to catastrophic equipment failures (e.g., compressor breakdown or vacuum leaks). In such scenarios, the true thermal dynamics will diverge irreversibly from the recipe, and the static prior may mislead the forecasting model, potentially leading to physically implausible predictions. A further limitation is that matching parameter counts across architectures does not perfectly equate training difficulty. We mitigated this by applying early stopping and uniform learning rate scheduling to all models.
While accurate, the fixed-weight strategy entails substantial hyperparameter optimization costs: for instance, PKAN\_fixed at 115,000 parameters required over 100,000 seconds ($\approx$28~hours) of training time on our hardware, rendering it impractical for rapid deployment or iterative process development. In such scenarios, the uncertainty-based or RBA 
strategies — which require only a single training run — are strongly recommended as computationally viable alternatives without significant accuracy penalties.
 
Future research should focus on three main directions to address these challenges and expand the framework's utility:
\begin{itemize}

    \item  Implementing dynamic anomaly detection modules — based, for instance, on statistical change-point detection or Bayesian online learning — capable of automatically reducing the 
    prior weight ($\lambda \to 0$) when real-time sensor data irreversibly diverges from the intended recipe. This would make 
    the PIF framework robust to catastrophic equipment failures such as compressor breakdowns or vacuum leaks, which currently represent its primary limitation.

    \item Extending the PIF architecture from 1-step-ahead direct forecasting to multi-step-ahead rollouts to directly support Model Predictive Control (MPC) strategies. This extension would enable the framework to act not only as a monitoring tool but as an active component of closed-loop process control in GxP-regulated manufacturing.

    \item Validating the PIF framework across different lyophilizer scales — from laboratory to pilot and full production scale — through systematic fine-tuning transfer learning experiments. Scale-up introduces substantially different heat and mass transfer dynamics, and demonstrating that process-informed representations remain transferable across equipment generations would significantly broaden the practical impact of this methodology.

\end{itemize}

\section*{Credit Authorship Contribution Statement}
\textbf{Ramona Rubini:} Writing - original draft, Methodology, Conceptualization, Visualization, Formal analysis, Software, Data curation \textbf{Siavash Khodakarami:} Methodology, Formal analysis, Visualization, \textbf{Aniruddha Bora:} Methodology, Formal analysis, Visualization, \textbf{George Em Karniadakis:} Writing – review and editing, Supervision, Resources, \textbf{Michele Dassisti:} Writing – review and editing, Supervision, Project administration, Funding acquisition

\section*{Data and Code Availability:}

The data and code will be made available after publication at: \href{https://github.com/ramonarubini/process-informed_forecasting.git}{GitHub Repository}

\section*{Acknowledgements}

This work was supported by the Italian Ministry of University and Research under the Programme “Department of Excellence” 
Legge 232/2016 (Grant No. CUP - D93C23000100001)”. This work was partially funded also by Sanofi Italia (Grant No. 
CUP - H91I22000400007). The authors acknowledge the computational resources and services at the Center for Computation and Visualization (CCV), Brown University.

\bibliographystyle{elsarticle-num}
\bibliography{bibliography}

\appendix
\section{Additional Experiments on the First Lyophilization Process}
\label{app2}

\begin{table}[H]
\centering
\caption{Performance comparison of seven deep learning model families and their PIF models, Fixed-weight (fixed), Uncertainty-based adaptive (uncertainty) and Residual-Based Attention (RBA), all configured to approximately 15,000 parameters for a fair comparison. The optimal models depends on the specific deep learning architecture. Uncertainty-based models and RBA-based models stand out for their computational efficiency providing a balanced performance profile. Standard models often have the lowest RMSE for architectures like MLP and Transformer, however the PI models have a better balance between accuracy, physical consistency and computational cost.}\label{tab:dl_models_15k}%
% \tiny
\resizebox{\textwidth}{!}{
\begin{tabular}{lrrrrrrr}
\hline
\textbf{Model Family} & \textbf{RMSE} & \textbf{$L_\infty Error$} & \textbf{GradError} & \textbf{$L_\infty(Grad)$} & \textbf{PVR (\%)} & \textbf{MO} & \textbf{Time(s)} \\
\hline
Recipe Prior Baseline & 0.02586 & 0.16179 & 0.00162 & 0.02100 & 0.00 & 0.000 & N/A \\
\hline
\multicolumn{8}{l}{\textbf{RNN Models}} \\
\hline
RNN\_fixed & \textbf{0.0067} & \textbf{0.0346} & 0.0019 & 0.0163 & 0.4782 & \textbf{0.0358} & 151.2 \\
RNN & 0.0074 & 0.0372 & \textbf{0.0018} & 0.0154 & 0.4782 & 0.0409 & 10.8 \\
RNN\_uncertainty & 0.0081 & 0.0371 & 0.0019 & 0.0153 & 0.4782 & 0.0395 & \textbf{5.8} \\
RNN\_RBA & 0.0109 & 0.0413 & 0.0019 & \textbf{0.0147} & \textbf{0.2657} & \textbf{0.0291} & 7.2 \\
\hline
\multicolumn{8}{l}{\textbf{cKAN Models}} \\
\hline
cKAN & \textbf{0.0087} & \textbf{0.0638} & 0.0017 & 0.0153 & 0.1063 & 0.0293 & 10.1 \\
cPKAN\_fixed & 0.0129 & 0.1172 & 0.0017 & 0.0149 & 0.3188 & 0.0213 & 119.0 \\
cPKAN\_uncertainty & 0.0148 & 0.0991 & 0.0017 & 0.0157 & \textbf{0.0000} & 0.0280 & \textbf{9.4} \\
cPKAN\_RBA & 0.0156 & 0.1004 & \textbf{0.0016} & \textbf{0.0144} & 0.0531 & \textbf{0.0183} & 15.9 \\
\hline
\multicolumn{8}{l}{\textbf{KAN Models}} \\
\hline
KAN & \textbf{0.0092} & \textbf{0.0501} & \textbf{0.0020} & \textbf{0.0202} & 0.3188 & 0.0259 & \textbf{969.3} \\
PKAN\_fixed & 0.0093 & 0.0502 & 0.0020 & 0.0203 & 0.3188 & 0.0259 & 14971.3 \\
PKAN\_uncertainty & 0.0145 & 0.1033 & 0.0215 & \textbf{0.0191} & \textbf{0.2657} & \textbf{0.0203} & 966.4 \\
PKAN\_RBA & 0.0155 & 0.1065 & 0.0207 & 0.0217 & 0.3188 & 0.0284 & 1179.0 \\
\hline
\multicolumn{8}{l}{\textbf{LEM Models}} \\
\hline
LEM\_fixed & \textbf{0.0095} & 0.0855 & 0.0019 & \textbf{0.0150} & \textbf{0.2657} & 0.0292 & 31632.1 \\
LEM & 0.0116 & 0.0957 & 0.0018 & 0.0156 & 0.4251 & \textbf{0.0234} & \textbf{2160.8} \\
LEM\_uncertainty & 0.0118 & 0.0702 & \textbf{0.0018} & 0.0162 & 0.3719 & 0.0359 & 1827.5 \\
LEM\_RBA & 0.0118 & \textbf{0.0607} & 0.0018 & 0.0153 & 0.3188 & 0.0321 & 2511.7 \\
\hline
\multicolumn{8}{l}{\textbf{LSTM Models}} \\
\hline
LSTM\_uncertainty & \textbf{0.0112} & \textbf{0.0655} & 0.0020 & \textbf{0.0151} & 0.3719 & 0.0378 & \textbf{12.3} \\
LSTM & 0.0116 & 0.0820 & \textbf{0.0020} & 0.0156 & 0.3719 & \textbf{0.0300} & 14.1 \\
LSTM\_fixed & 0.0118 & 0.0656 & 0.0020 & 0.0157 & \textbf{0.3188} & 0.0405 & 161.3 \\
LSTM\_RBA & 0.0135 & 0.0676 & 0.0020 & 0.0153 & 0.3188 & 0.0312 & 20.5 \\
\hline
\multicolumn{8}{l}{\textbf{MLP Models}} \\
\hline
MLP\_fixed & \textbf{0.0117} & \textbf{0.0707} & 0.0018 & 0.0186 & \textbf{0.0000} & 0.0275 & 119.5 \\
MLP & 0.0125 & 0.1016 & \textbf{0.0017} & 0.0203 & 0.1063 & 0.0295 & \textbf{8.3} \\
MLP\_uncertainty & 0.0150 & 0.0878 & 0.0018 & \textbf{0.0182} & 0.0000 & 0.0181 & 8.7 \\
MLP\_RBA & 0.0159 & 0.0929 & 0.0018 & 0.0189 & 0.0000 & \textbf{0.0146} & 8.7 \\
\hline
\multicolumn{8}{l}{\textbf{Transformer Models}} \\
\hline
Transformer & \textbf{0.0144} & \textbf{0.0789} & \textbf{0.0015} & 0.0519 & \textbf{0.0000} & 0.0292 & 59.6 \\
Transformer\_uncertainty & 0.0241 & 0.1051 & 0.0016 & \textbf{0.0386} & 0.3719 & \textbf{0.0163} & \textbf{34.3} \\
Transformer\_RBA & 0.0242 & 0.1453 & 0.0016 & 0.0400 & 0.1063 & 0.0438 & 67.4 \\
Transformer\_fixed & 0.0507 & 0.1686 & 0.0015 & 0.0505 & 0.0000 & 0.0417 & 702.6 \\
\hline
\end{tabular}}
\end{table}

\begin{table}[H]
\centering
\caption{Performance metrics for deep learning models and their PIF models with model complexity increased to approximately 45,000 parameters. At this increased model capacity, process-informed models outperform their standard data-driven counterparts in predictive accuracy across almost all architectures. The Fixed-weight (fixed) and Uncertainty-based (uncertainty) methods are the top performers, with the lowest RMSE for multiple models like LEM, cKAN and LSTM. The uncertainty-based models are the best balance, often the best or near best and one of the most computationally efficient. This suggests that as model capacity grows, the integration of process knowledge becomes increasingly beneficial, with adaptive methods like uncertainty-based offering the most compelling combination of accuracy and practical scalability.}\label{tab:dl_models_45k}%
% \tiny
\resizebox{\textwidth}{!}{
\begin{tabular}{lrrrrrrr}
\hline
\textbf{Model Family} & \textbf{RMSE} & \textbf{$L_\infty Error$} & \textbf{GradError} & \textbf{$L_\infty(Grad)$} & \textbf{PVR (\%)} & \textbf{MO} & \textbf{Time(s)} \\
\hline
Recipe Prior Baseline & 0.02586 & 0.16179 & 0.00162 & 0.02100 & 0.00 & 0.000 & N/A \\
\hline
\multicolumn{8}{l}{\textbf{RNN Models}} \\
\hline
RNN & \textbf{0.0066} & \textbf{0.0299} & 0.0018 & 0.0167 & 0.6376 & 0.0354 & 13.1 \\
RNN\_fixed & 0.0068 & 0.0313 & \textbf{0.0018} & 0.0153 & 0.5313 & 0.0318 & 169.9 \\
RNN\_RBA & 0.0087 & 0.0428 & 0.0018 & \textbf{0.0152} & \textbf{0.4782} & 0.0395 & 8.5 \\
RNN\_uncertainty & 0.0089 & 0.0468 & 0.0018 & 0.0166 & 0.5845 & \textbf{0.0289} & \textbf{8.0} \\
\hline
\multicolumn{8}{l}{\textbf{cKAN Models}} \\
\hline
cPKAN\_fixed & \textbf{0.0092} & \textbf{0.0799} & \textbf{0.0016} & \textbf{0.0143} & 0.1594 & 0.0292 & 118.4 \\
cKAN & 0.0134 & 0.1111 & 0.0017 & 0.0174 & 0.3188 & 0.0321 & 10.5 \\
cPKAN\_RBA & 0.0166 & 0.0940 & 0.0016 & 0.0143 & \textbf{0.0531} & 0.0169 & 8.3 \\
cPKAN\_uncertainty & 0.0225 & 0.0997 & 0.0017 & 0.0166 & 0.2657 & \textbf{0.0122} & \textbf{4.8} \\
\hline
\multicolumn{8}{l}{\textbf{KAN Models}} \\
\hline
KAN & \textbf{0.0079} & \textbf{0.0435} & 0.0020 & \textbf{0.0160} & 0.5313 & 0.0196 & \textbf{2970.3} \\
PKAN\_fixed & 0.0081 & 0.0451 & 0.0020 & 0.0160 & 0.5313 & \textbf{0.0195} & 47535.0 \\
PKAN\_RBA & 0.0151 & 0.1006 & 0.0021 & 0.0182 & 0.4251 & 0.0209 & 3185.7 \\
PKAN\_uncertainty & 0.0152 & 0.0987 & \textbf{0.0020} & 0.0171 & \textbf{0.3719} & 0.0232 & 3008.4 \\
\hline
\multicolumn{8}{l}{\textbf{LEM Models}} \\
\hline
LEM\_fixed & \textbf{0.0083} & 0.0697 & 0.0018 & \textbf{0.0155} & 0.4251 & 0.0280 & 34187.3 \\
LEM & 0.0098 & 0.0867 & 0.0018 & 0.0159 & \textbf{0.2657} & 0.0285 & 2379.8 \\
LEM\_uncertainty & 0.0107 & 0.0615 & \textbf{0.0018} & 0.0162 & 0.2657 & 0.0370 & \textbf{1717.2} \\
LEM\_RBA & 0.0117 & \textbf{0.0596} & 0.0019 & 0.0162 & 0.3188 & \textbf{0.0261} & 2878.2 \\
\hline
\multicolumn{8}{l}{\textbf{LSTM Models}} \\
\hline
LSTM\_fixed & \textbf{0.0098} & 0.0724 & 0.0020 & 0.0155 & 0.3719 & 0.0370 & 196.8 \\
LSTM & 0.0099 & 0.0675 & \textbf{0.0020} & 0.0156 & \textbf{0.3188} & \textbf{0.0353} & 14.7 \\
LSTM\_uncertainty & 0.0111 & \textbf{0.0547} & 0.0020 & \textbf{0.0154} & 0.3719 & 0.0340 & 14.7 \\
LSTM\_RBA & 0.0130 & 0.0657 & 0.0020 & 0.0155 & 0.3719 & 0.0395 & \textbf{14.5} \\
\hline
\multicolumn{8}{l}{\textbf{MLP Models}} \\
\hline
MLP\_fixed & \textbf{0.0089} & \textbf{0.0523} & 0.0018 & 0.0177 & 0.0531 & 0.0237 & 105.3 \\
MLP & 0.0105 & 0.0821 & \textbf{0.0017} & 0.0183 & 0.1063 & 0.0180 & 8.2 \\
MLP\_uncertainty & 0.0153 & 0.0929 & 0.0018 & 0.0189 & \textbf{0.0000} & 0.0194 & 8.9 \\
MLP\_RBA & 0.0165 & 0.0990 & 0.0017 & \textbf{0.0170} & 0.0000 & \textbf{0.0105} & \textbf{6.5} \\
\hline
\multicolumn{8}{l}{\textbf{Transformer Models}} \\
\hline
Transformer\_fixed & \textbf{0.0130} & 0.0869 & 0.0016 & 0.0496 & 0.0531 & \textbf{0.0120} & 779.8 \\
Transformer\_uncertainty & 0.0145 & 0.0995 & 0.0015 & 0.0430 & 0.0531 & 0.0302 & \textbf{40.5} \\
Transformer & 0.0150 & \textbf{0.0693} & \textbf{0.0015} & 0.0453 & 0.1594 & 0.0152 & 48.4 \\
Transformer\_RBA & 0.0160 & 0.0786 & 0.0015 & \textbf{0.0150} & \textbf{0.0000} & 0.0293 & 81.2 \\
\hline
\end{tabular}}
\end{table}

\section{Validation on a Secondary Lyophilization Process}
\label{app3}

To assess the generalizability of our findings, we conducted a validation study on a secondary unseen lyophilization dataset. This experiment demonstrates the robustness of our conclusions and evaluates the potential of the top-performing models to adapt to new process dynamics. The analysis focuses on the deep learning models at the $\sim$ 30,000 parameter scale, as they are the most capable architectures from our primary study. The secondary dataset is sourced from a lyophilization cycle for a different biopharmaceutical product with different thermal dynamics, different ramp rates and phase durations. The data are preprocessed and partitioned using the same methodology as the first dataset to ensure a consistent evaluation framework. All models are retrained from scratch on the new training set and evaluated on the test set. 

The results are summarized in Table \ref{tab:evaluation_metrics_grouped_rmse_2}. The KAN and cKAN families are still state of the art, with the lowest RMSE and show strong function approximation capabilities on new data but also strong adherence to thermodynamic safety, as evidenced by their minimal Maximum Overshoot (MO) values. The fixed-weight  models are competitive, closely following their pure counterparts in accuracy and physical consistency maintaining near-zero PVR during the extended hold phases of this new product. On this secondary dataset, the Transformer architecture and its fixed-weight variant have the lowest value of Gradient Error, and low PVR. Thus, they are better at capturing the underlying temporal structure and trends of this process, even if their overall point-wise accuracy is lower. A significant finding from this validation study is the performance of the uncertainty-weighted models. These variants are competitive on the source dataset and consistently rank lower on this new process. This means that while the uncertainty-based weighting scheme is effective at optimizing the data-physics trade-off for a specific process, it may be less robust to domain shifts compared to the simpler fixed-weight process-informed integration. However, despite any relative drop in point-wise accuracy, these PIF variants maintain lower PVR and MO compared to pure data-driven baselines. The key findings from this validation study are that while the MLP architecture is more robust on this specific process effectively capping the Maximum Overshoot under severe sensor perturbation, the overall class of robust models (MLP, KAN-family) remains consistent and the fixed-weight integration shows more stable performance across domains than the uncertainty-based method. 

\begin{table}[H]
\centering
\caption{Performance evaluation of deep learning models and their PIF variants on the secondary lyophilization dataset, with model complexity increased to approximately 30,000 parameters. The table compares Fixed-weight (\_fixed), Uncertainty-based (\_uncertainty), and Residual-Based Attention (\_RBA) integration strategies against their pure data-driven counterparts.  The KAN-family architectures achieve the lowest RMSE values with cPKAN\_Uncertainty model representing the most computationally efficient solution.
}\label{tab:evaluation_metrics_grouped_rmse_2}%
% \tiny
\resizebox{\textwidth}{!}{
\begin{tabular}{lrrrrrrr}
\hline
\textbf{Model Family} & \textbf{RMSE} & \textbf{$L_\infty Error$} & \textbf{GradError} & \textbf{$L_\infty(Grad)$} & \textbf{PVR (\%)} & \textbf{MO} & \textbf{Time(s)} \\
\hline
Recipe Prior Baseline & 0.0116 & 0.1260 & 0.0021 & 0.0485 & 0.00 & 0.0000 & N/A \\
\hline
\multicolumn{8}{l}{\textbf{RNN Models}} \\
\hline
RNN\_RBA & \textbf{0.0093} & \textbf{0.0962} & 0.0031 & 0.0494 & 1.1127 & \textbf{0.1976} & 18.1 \\
RNN\_uncertainty & 0.0096 & 0.1155 & \textbf{0.0027} & 0.0531 & \textbf{0.7881} & 0.2058 & \textbf{16.5} \\
RNN & 0.0106 & 0.1232 & 0.0026 & 0.0496 & 0.7418 & 0.2142 & 14.5 \\
RNN\_fixed & 0.0112 & 0.1018 & 0.0032 & \textbf{0.0438} & 0.9736 & 0.2022 & 247.7 \\
\hline
\multicolumn{8}{l}{\textbf{cKAN Models}} \\
\hline
cPKAN\_uncertainty & \textbf{0.0090} & \textbf{0.1015} & 0.0023 & \textbf{0.0361} & 0.1391 & 0.2028 & 12.4 \\
cKAN & 0.0103 & 0.1257 & 0.0022 & 0.0494 & 0.2318 & 0.2211 & 16.9 \\
cPKAN\_fixed & 0.0110 & 0.1222 & \textbf{0.0022} & 0.0489 & 0.2782 & 0.2176 & 153.0 \\
cPKAN\_RBA & 0.0119 & 0.1083 & 0.0022 & 0.0488 & \textbf{0.1391} & \textbf{0.2036} & \textbf{10.2} \\
\hline
\multicolumn{8}{l}{\textbf{KAN Models}} \\
\hline
PKAN\_fixed & \textbf{0.0082} & 0.1019 & \textbf{0.0023} & \textbf{0.0356} & 0.1854 & 0.2019 & 36075.5 \\
PKAN\_uncertainty & 0.0089 & 0.0919 & 0.0024 & 0.0419 & 0.0927 & 0.1932 & \textbf{2304.3} \\
PKAN\_RBA & 0.0092 & \textbf{0.0836} & 0.0024 & 0.0414 & \textbf{0.0464} & \textbf{0.1849} & 2800.1 \\
\hline
\multicolumn{8}{l}{\textbf{LEM Models}} \\
\hline
LEM\_RBA & \textbf{0.0110} & 0.1205 & 0.0031 & \textbf{0.0524} & 0.8345 & \textbf{0.2158} & 3523.9 \\
LEM\_uncertainty & 0.0111 & \textbf{0.1196} & \textbf{0.0029} & 0.0565 & \textbf{0.7881} & 0.2167 & \textbf{1919.2} \\
LEM\_fixed & 0.0153 & 0.1873 & 0.0027 & 0.0541 & 1.0199 & 0.2650 & 62601.0 \\
LEM & 0.0154 & 0.1923 & 0.0027 & 0.0548 & 1.0199 & 0.2734 & 3264.1 \\
\hline
\multicolumn{8}{l}{\textbf{LSTM Models}} \\
\hline
LSTM\_uncertainty & \textbf{0.0110} & \textbf{0.1279} & 0.0031 & \textbf{0.0541} & \textbf{0.8345} & \textbf{0.2233} & \textbf{18.7} \\
LSTM\_RBA & 0.0124 & 0.1490 & 0.0029 & 0.0590 & 1.0663 & 0.2435 & 18.9 \\
LSTM\_fixed & 0.0126 & 0.1617 & 0.0030 & 0.0621 & 0.8345 & 0.2562 & 275.6 \\
LSTM & 0.0146 & 0.1853 & \textbf{0.0027} & 0.0560 & 1.2054 & 0.2690 & 27.8 \\
\hline
\multicolumn{8}{l}{\textbf{MLP Models}} \\
\hline
MLP\_uncertainty & \textbf{0.0118} & \textbf{0.1439} & \textbf{0.0023} & 0.0559 & \textbf{0.2782} & 0.2394 & \textbf{16.2} \\
MLP & 0.0125 & 0.1594 & 0.0023 & \textbf{0.0485} & 0.5100 & 0.2429 & 16.8 \\
MLP\_fixed & 0.0128 & 0.1571 & 0.0024 & 0.0504 & 0.4172 & \textbf{0.2385} & 225.0 \\
MLP\_RBA & 0.0134 & 0.1675 & 0.0023 & 0.0523 & 0.4172 & 0.2455 & 14.2 \\
\hline
\multicolumn{8}{l}{\textbf{Transformer Models}} \\
\hline
Transformer\_uncertainty & \textbf{0.0136} & \textbf{0.1236} & \textbf{0.0020} & \textbf{0.0369} & \textbf{0.1391} & \textbf{0.2266} & \textbf{65.7} \\
Transformer\_RBA & 0.0162 & 0.1449 & 0.0020 & 0.0717 & 0.1391 & 0.2405 & 79.8 \\
Transformer & 0.0181 & 0.1559 & 0.0021 & 0.0678 & 0.1391 & 0.2512 & 67.7 \\
\hline
\end{tabular}}
\end{table}

\begin{figure}[H]
    \centering
    \includegraphics[width=\textwidth]{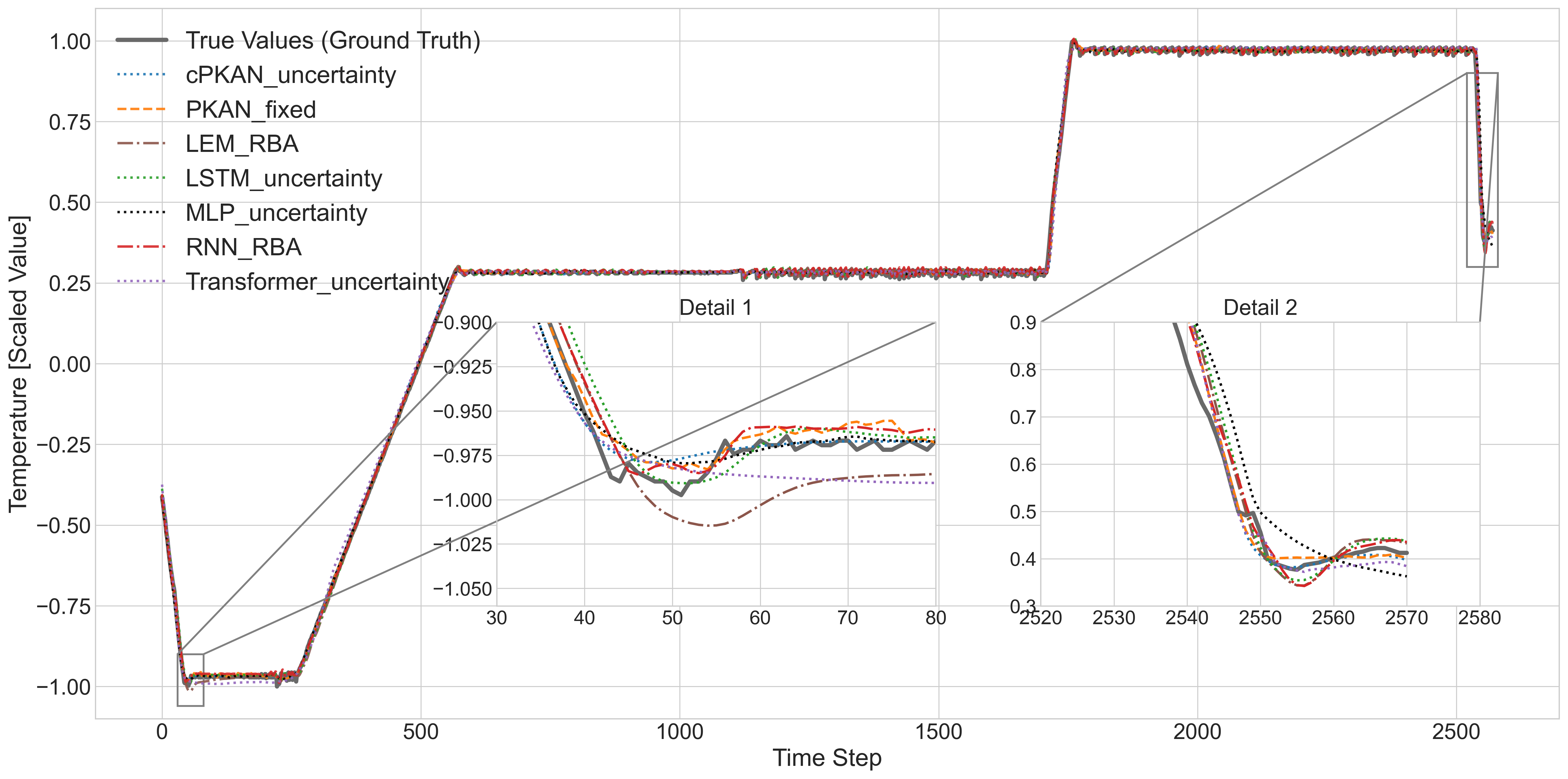}
    \caption{Comparative analysis of model predictions for the thermal dynamics of the secondary lyophilization process. The figure shows the output of the best-performing variant from seven different model families ($\sim{30,000}$ parameters). Compared to the primary dataset, all architectures generalize well to this novel thermal profile, confirming the robustness of the PIF methodology across different pharmaceutical products.}
    \label{fig:30kpred2dataset}
\end{figure}

\begin{figure}[H]
    \centering
    \includegraphics[width=\textwidth]{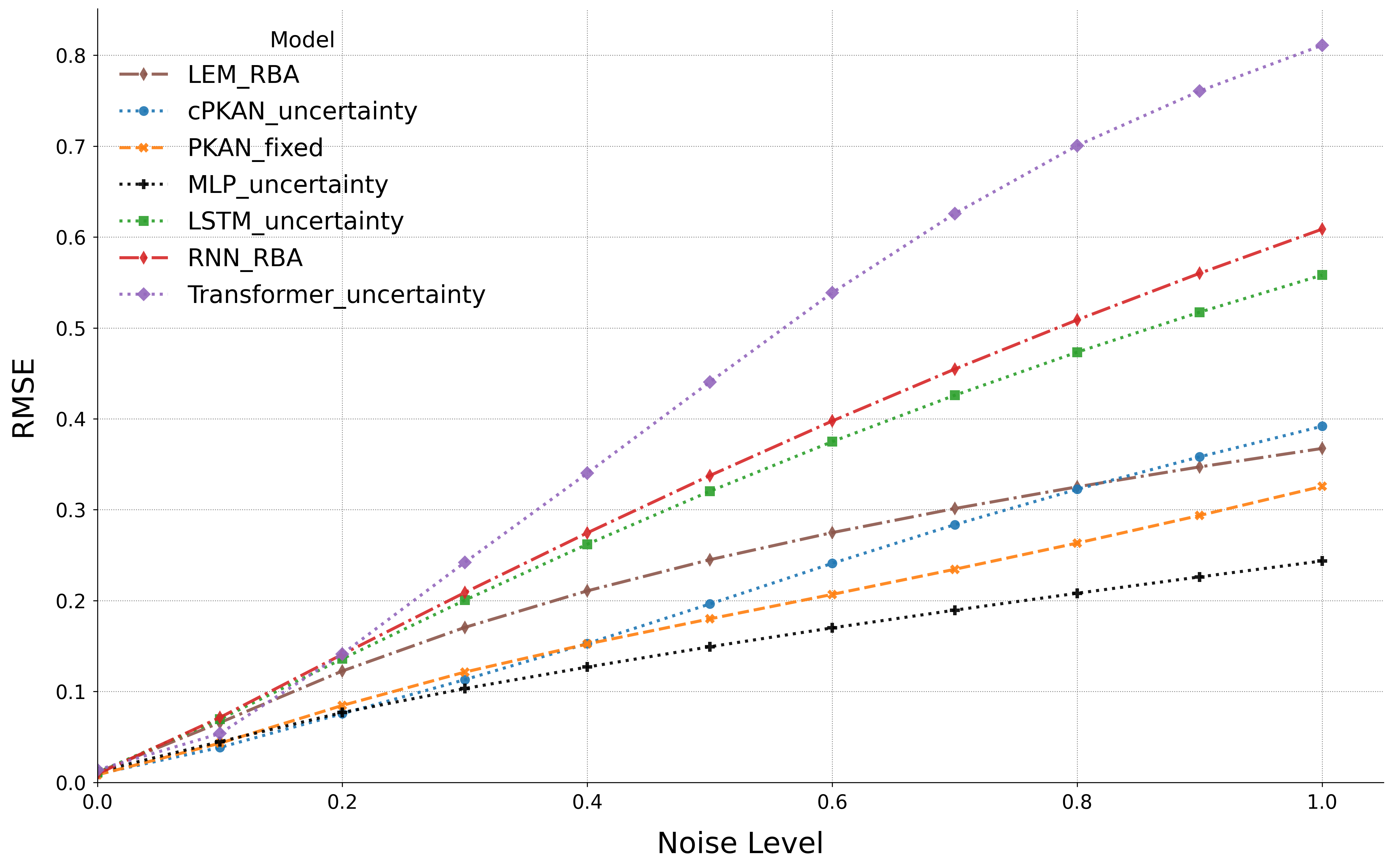}
    \caption{Robustness evaluation of models of approximately 30,000 parameters on a secondary lyophilization process. This chart evaluates how the predictive accuracy of different models deteriorates as the input data becomes noisier. The results identify the MLP architecture as the most resilient to such perturbations, maintaining the lowest error across all noise levels. The PKAN\_fixed model and LEM model also demonstrate strong robustness.  }
    \label{fig:robust_input_2dataset}
\end{figure}

\begin{table}[H]
\centering
\caption{Detailed performance metrics at the 0.60 input noise level for the secondary lyophilization process. This table provides a quantitative breakdown of performance under severe sensor perturbation, evaluating prediction accuracy (RMSE), system dynamics (Gradient Error), and physical safety (PVR and Max Overshoot). The results highlight a performance trade-off: while the MLP architecture is the most robust in terms of overall point-wise prediction accuracy (lowest RMSE), the cKAN model excels in capturing the instantaneous rate of change (lowest Gradient Error) and maintaining thermodynamic safety. The cKAN\_Uncertainty model achieves the lowest Physical Violation Rate (81.87\%) and the absolute minimum Max Overshoot (0.3838), preventing dangerous temperature spikes that other models fail to suppress.} \label{tab:noise_input_0600_2dataset}
\resizebox{\textwidth}{!}{
\begin{tabular}{lrrrrrr}
\hline
\multicolumn{7}{c}{\textbf{Noise Level: 0.60}} \\
\hline
\textbf{Model} & \textbf{RMSE} & \textbf{$L_\infty$Error} & \textbf{GradError} & \textbf{$L_\infty$(Grad)} & \textbf{PVR (\%)} & \textbf{MO} \\
\hline
MLP\_uncertainty & \textbf{0.1699} & \textbf{0.6262} & 0.0888 & 0.5267 & 91.33 & 0.6124 \\
PKAN\_fixed & 0.2068 & 0.7682 & 0.0939 & 0.5006 & 93.18 & 0.6973 \\
cPKAN\_uncertainty & 0.2409 & 0.7849 & \textbf{0.0619} & \textbf{0.4215} & \textbf{81.87} & \textbf{0.3838} \\
LEM\_RBA & 0.2748 & 1.0001 & 0.1461 & 0.7023 & 94.90 & 0.9532 \\
LSTM\_uncertainty & 0.3749 & 1.4683 & 0.2031 & 0.8985 & 97.21 & 1.3634 \\
RNN\_RBA & 0.3976 & 1.4615 & 0.2237 & 1.0352 & 97.07 & 1.4621 \\
Transformer\_uncertainty & 0.5387 & 1.9235 & 0.2610 & 0.9846 & 70.65 & 1.7930 \\
\hline
\end{tabular}
}
\end{table}

\begin{figure}[H]
    \centering
    \includegraphics[width=\textwidth]{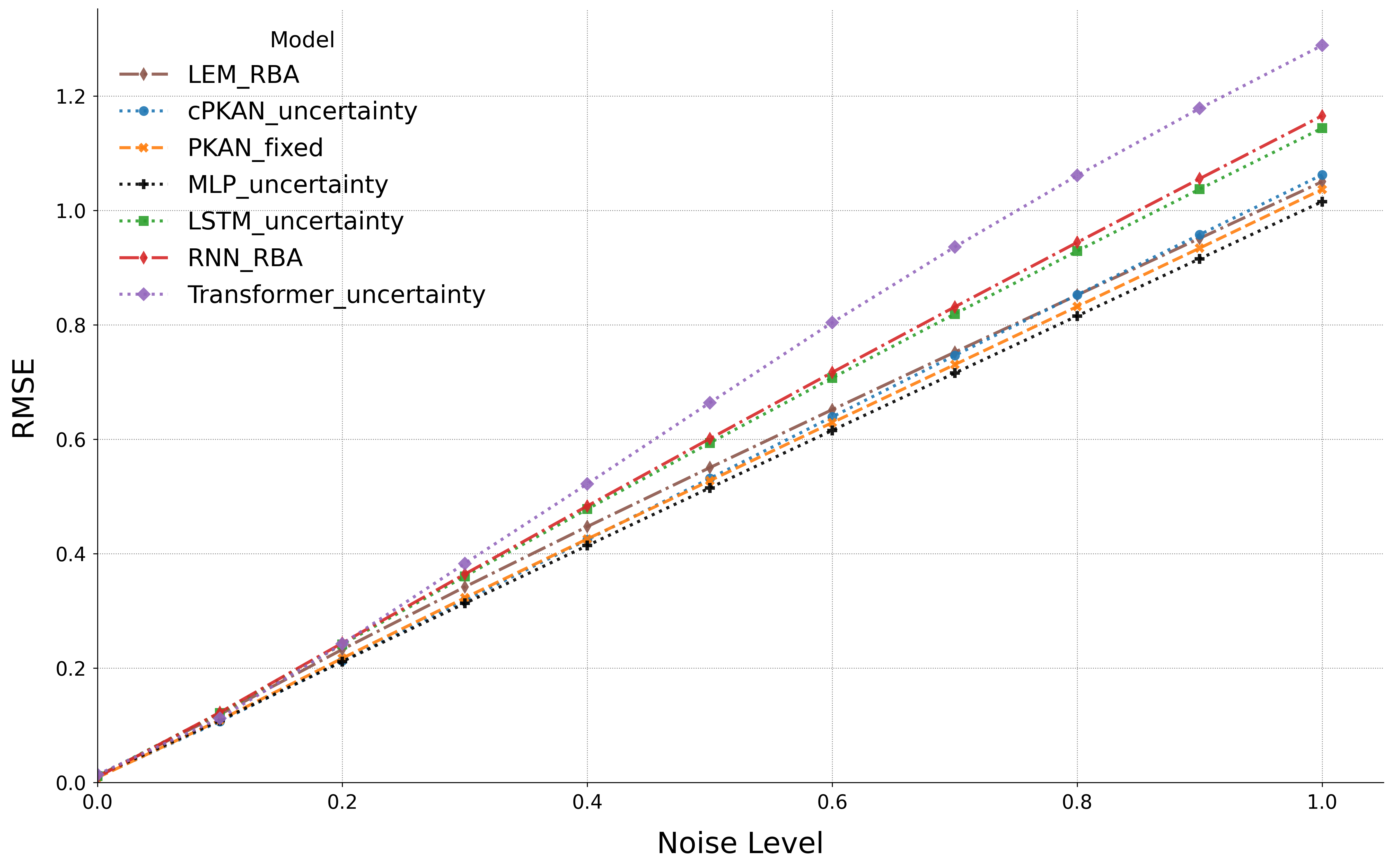}
    \caption{Robustness evaluation of model of approximately 30,000 parameters on a secondary lyophilization process. This chart evaluates the results of the system noise robustness test, where both input features (X) and the target variable (y) are corrupted with independent, zero-mean Gaussian noise. While the relative performance ranking is consistent with the input noise analysis, with the MLP being the most robust model, the overall performance of all models degrades substantially and more uniformly. This convergence of performance under system-wide noise is consistent with the findings on the primary dataset.}
    \label{fig:robust_system_2dataset}
\end{figure}

\begin{figure}[H]
    \centering
    \includegraphics[width=\textwidth]{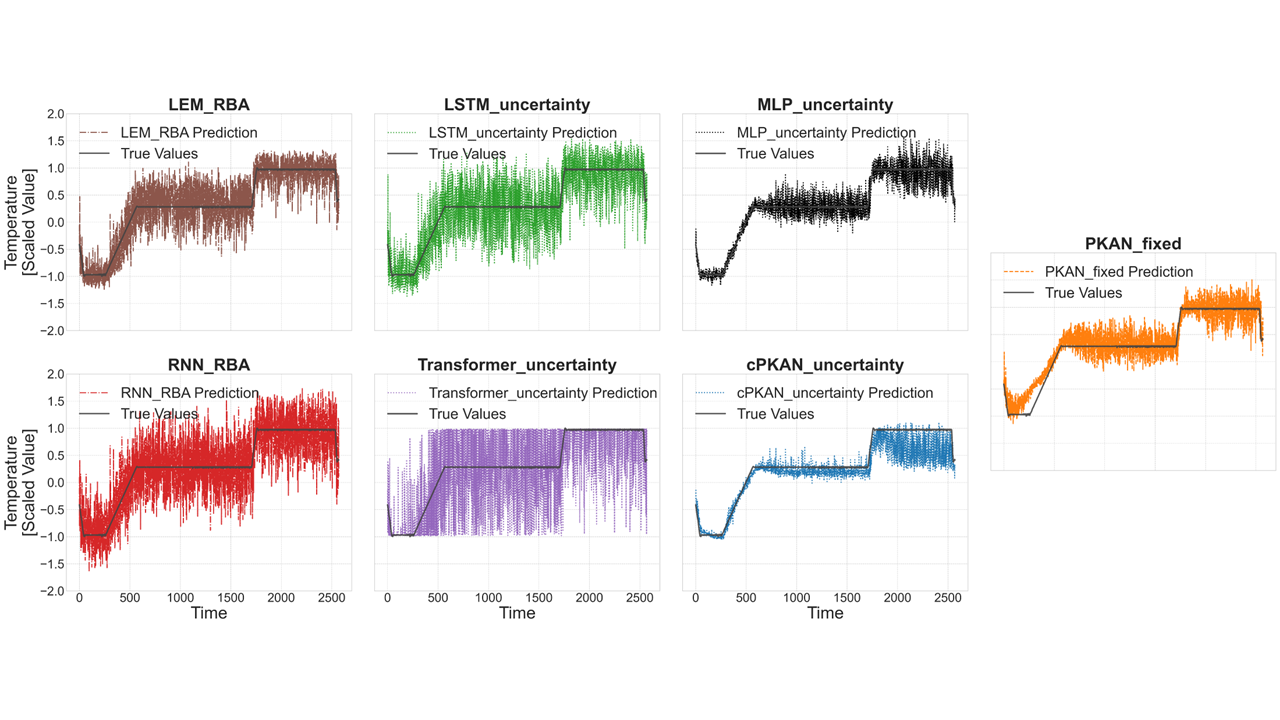}
    \caption{Model prediction comparison on noisy data (Noise $\sigma=0.6$).This figure shows the prediction of the best-performing models from each family ($\sim30,000$) parameters on a test set where the input features are corrupted with Gaussian Noise ($\sigma=0.6$). The predictions are plotted against the original, noise-free ground truth to assess noise-filtering capability.The cPKAN\_Uncertainty model and the MLP\_Uncertainty model demonstrate the strongest noise-filtering capability, producing a smooth trajectory that closely follows the ground truth despite the severe sensor perturbation. In contrast, RNN\_RBA and Transformer\_Uncertainty exhibit high prediction variance throughout the cycle, making them unsuitable for monitoring tasks under noisy conditions.}
    \label{fig:robust0.60_2dataset}
\end{figure}

\end{document}